\documentclass{article}

\usepackage[dblblindworkshop,final]{neurips_2025}

\usepackage{float}          
\usepackage{graphicx}       
\graphicspath{ {./images/} }
\usepackage[utf8]{inputenc} 
\usepackage[T1]{fontenc}    
\usepackage{hyperref}       
\usepackage{url}            
\usepackage{booktabs}       
\usepackage{amsfonts}       
\usepackage{nicefrac}       
\usepackage{microtype}      
\usepackage{xcolor}         
\usepackage{pdflscape}      
\usepackage{booktabs}
\usepackage{threeparttable}
\usepackage{amsmath}       
\usepackage{longtable}
\usepackage{caption}
\usepackage{subcaption}

\title{Large Language Models as Medical Codes Selectors: a benchmark using the International Classification of Primary Care}
\workshoptitle{The Second Workshop on GenAI for Health: Potential, Trust, and Policy Compliance}

\author{
  Vinicius Anjos de Almeida\thanks{Medical School, University of São Paulo, São Paulo, Brazil. ORCID: 0009-0001-1273-586X. Contact: \texttt{vinicius.almeida@alumni.usp.br}}  \and
  Vinicius de Camargo\thanks{Department of Epidemiology, School of Public Health, University of São Paulo, São Paulo, Brazil. ORCID: 0009-0004-9885-9573} \and
  Raquel Gómez-Bravo\thanks{Rehaklinik, Centre Hospitalier Neuro-psychiatrique (CHNP), Ettelbruck, Luxembourg. ORCID: 0000-0002-3786-8626} \and
  Kees van Boven\thanks{Department of Primary and Community Care, Radboud University, Nijmegen, Netherlands. ORCID: 0000-0002-4422-9443} \and
  Egbert van der Haring\thanks{Independant researcher, Netherlands} \and
  Marcelo Finger\thanks{Department of Computer Science, Institute of Mathematics and Statistics, University of Sao Paulo, São Paulo, Brazil. ORCID: 0000-0002-1391-1175} \and
  Luis Fernandez Lopez\thanks{Medical School, University of São Paulo, São Paulo, Brazil. ORCID: 0000-0003-
4097-6887}
}

\begin{document}

\maketitle

\begin{abstract}
    \textbf{Background}: Medical coding structures healthcare data for research, quality monitoring, and policy. This study assesses the potential of large language models (LLMs) to assign ICPC-2 codes using the output of a domain-specific search engine. \\
    \textbf{Methods}: A dataset of 437 Brazilian Portuguese clinical expressions, each annotated with ICPC-2 codes, was used. A semantic search engine (OpenAI's text-embedding-3-large) retrieved candidates from 73,563 labeled concepts. Thirty-three LLMs were prompted with each query and retrieved results to select the best-matching ICPC-2 code. Performance was evaluated using F1-score, along with token usage, cost, response time, and format adherence. \\
    \textbf{Results}: Twenty-eight models achieved F1-score > 0.8; ten exceeded 0.85. Top performers included gpt-4.5-preview, o3, and gemini-2.5-pro. Retriever optimization can improve performance by up to 4 points. Most models returned valid codes in the expected format, with reduced hallucinations. Smaller models (<3B) struggled with formatting and input length. \\
    \textbf{Conclusions}: LLMs show strong potential for automating ICPC-2 coding, even without fine-tuning. This work offers a benchmark and highlights challenges, but findings are limited by dataset scope and setup. Broader, multilingual, end-to-end evaluations are needed for clinical validation. \\
  
\end{abstract}

\section{Introduction}

Medical coding organizes patient data, guides management and billing, and supports research. It is difficult, time-consuming, and often results in low-quality data \cite{Wockenfuss2009, Horsky2017-si}. Local or global coding practice changes can distort epidemiological data and impact decision-making \cite{Atkin2022, Lindenauer2012}. Automating medical coding can bring value through higher-quality real-world data in healthcare services and systems, foster research, and guide evidence-based policy-making.

\subsection{Previous work}
Medical coding is an extreme multi-label classification (XMC) task, defined as a multi-label classification problem where the number of categories is very large and data is often unbalanced \cite{dasgupta2025reviewextrememultilabelclassification}. As an example, the International Classification of Diseases, 10th edition, has more than 14,000 classes \cite{whoInternationalClassification}, excluding its even bigger expansions such as ICD-10-CM \cite{cdcICD10CM}.
Various techniques have been used to address similar problems. Two studies are specifically more related to this research project. 

D'Oosterlinck et al. \cite{doosterlinck2024incontextlearningextrememultilabel} proposed a general stepwise approach to label selection called Infer-Retrieve-Rank. First, a large language model (LLM) generates a set of candidate terms. Second, a retriever returns an ordered list of candidate labels. Finally, an LM re-ranks the retrieved labels. In this case, optimizations are made by selecting the underlying LMs and through structured few-shot prompt optimization. Although not applied to medical coding, their system outperformed specialized models in three XMC benchmarks. Zhu and Zamani \cite{ZhuZamani2023} focused on the problem of new labels or unknown relations between instances and labels. They framed the problem as zero-shot XMC. First, a shortlist of candidate labels was generated, supported by generated or retrieved examples. Then, the shortlists are mapped back to real labels, re-ranked, and the final prediction consists of the top-k labels.

Both methods have proven effective in experimental settings, but large-scale adoption still requires validation on real-world data. The initial successes of related systems suggest that implementation is feasible, provided they are complemented by hybridization techniques and human oversight.

\subsection{Objectives}
This research frames automated medical coding as an extract-retrieve-select task, with independently optimizable steps. This study focuses on the code selection step, specifically in the following research questions: Can large language models accurately choose ICPC-2 codes when given clinical expressions and search results? How do they compare with respect to price, token usage, number of parameters, answer formatting, or length of the search results list? How much can the performance of those models be improved by only optimizing the search engine?

\section{Methods}
The extract-retrieve-select framework consists of three steps: (1) extraction — extracting relevant concepts from the input (e.g., clinical note); (2) retrieval — retrieving candidate codes from a reference \textit{corpus}; and (3) selection — one of the retrieved codes \textemdash or none \textemdash as the predicted code for the given expression. Evaluating the extraction step is out of the scope of this study, and the retrieval step was already explored in previous work \cite{JMAI9766}. To properly evaluate the code selection step, the following components are required: a target clinical classification system (e.g., ICD-10); a \textit{corpus} containing a mapping between common expressions and their respective codes; a search engine with a fixed ranking algorithm; an evaluation dataset with real-world expressions and their mapped codes; and a set of large language models to be tested. 

\subsection{International Classification of Primary Care}

As the classification system, the International Classification of Primary Care, 2nd edition (ICPC-2) was selected. ICPC-2 is widely used in primary care across the globe. It covers common conditions, undifferentiated symptoms, procedures, and non-disease-related issues. In comparison with other systems, such as the International Classification of Diseases (ICD), it is concise, suitable for primary care settings globally, with about 1,300 different categories, and groups in a meaningful way the most prevalent clinical conditions. 

\subsection{ICPC-2 thesaurus}

The ICPC-2 thesaurus \cite{Becker2005ICPC2ICD10} was developed by the Ministry of Health of Belgium in collaboration with a group of researchers from the Family Medicine Department at the University of Amsterdam. The Brazilian Portuguese translation \cite{thesaurus}, published by the Brazilian Society of Family and Community Medicine in partnership with the Ministry of Health of Brazil, was used as the \textit{corpus} of the search engine. 

The thesaurus consists of a mapping between clinical concepts and ICPC-2 and ICD-10 codes. It was reorganized to build the search engine in the same way as Almeida \textit{et al.} \cite{JMAI9766, JMAI9766-source-code} did on their evaluation of different ranking algorithms for medical coding. The final \textit{corpus} contains 73,563 clinical concepts mapped to their respective ICPC-2 codes. Those concepts include technical terms, colloquial terms, acronyms, codes, and code titles. 

\subsection{Search engine}

Almeida \textit{et al.} \cite{JMAI9766} compared different ranking algorithms in the context of medical coding with ICPC-2 and found that semantic search with OpenAI's model \textit{text-3-embedding-large} \cite{openaiembedding} had the best performance in most ranking metrics. In this study, the search engine was reproduced using the same methodology for consistency. As the vector database, the Chroma DB \cite{trychromaChroma} was used in conjunction with the Hierarchical Navigable Small World algorithm to perform the similarity search. The embeddings were generated with OpenAI's \textit{text-3-embedding-large} model for each concept in the \textit{corpus}. 

\subsection{Baselines}

Two experiments will be used as baselines to compare with LLMs' performance: 1. the automatic selection of the first result of the search engine; and 2. the gpt-4o model prompted without access to the search engine results. 

\subsection{Evaluation dataset}

A dataset consisting of 437 clinical concepts written in Brazilian Portuguese annotated with relevant ICPC-2 codes was used as the evaluation dataset \cite{JMAI9766}. This dataset was organized with real-world queries from an ICPC-2 search engine. Each query was independently annotated by peers with experience in medical coding with the ICPC-2. The frequency of each ICPC-2 in the evaluation dataset is presented as a heatmap in the appendix \ref{appendix} as figures \ref{fig:heatmap_1} and \ref{fig:heatmap_2}.

\subsection{Large language models}

Various large language models (LLMs) were evaluated on the task of selecting an ICPC-2 code given a query and a list of search engine results. The models were chosen based on their popularity at the time of the study, diversity in size, presence or absence of reasoning capabilities, and a balance between open- and closed-source implementations. The list of the selected open- and closed-source models is presented in the tables \ref{tab:listofllmsopen} and \ref{tab:listofllmsclosed}. 

Inference was performed with different methods. Smaller open-source models ran on a local device described in detail in section \ref{section:hardware}. Bigger and private models' responses were obtained through third-party APIs from various companies, including OpenAI \cite{openaiOpenAIPlatform}, Google \cite{googleStructuredOutput}, HuggingFace \cite{huggingfaceHuggingFace}, and FireworksAI \cite{fireworksFireworksFastest}. The correspondence between models and hosting platforms is available in tables \ref{tab:listofllmsopen} and \ref{tab:listofllmsclosed}. Fine-tuning LLMs for automatic code selection was beyond the scope of this study.

The same prompt was used to interact with every model. Although our benchmark refers to code clinical concepts written in Brazilian Portuguese, the prompt was written in English, since it is commonly the most represented language in language models' pretraining data and mixed-language prompts seem to perform better than non-English prompts \cite{mondshine2025englishimpactprompttranslation}. 

The prompt includes two parameters: the query, which corresponds to the searched clinical concept; and the search results, which is a JSON string \cite{json} containing a ranked list of objects with the retrieved expression and an ICPC-2 code related to that retrieved expression. The JSON format was chosen for being a universal standard in digital information exchange and for being widely represented in large language models' pretraining data \cite{openaiOpenAIPlatform, llamaLlamaModel, googleStructuredOutput}. The prompt template is available in the table \ref{tab:prompt}. A complete example is available in the appendix \ref{appendix} in the table \ref{tab:promptexample}.

\begin{table}
    \centering
    \caption{Prompt template used to interact with every large language model.}
    \begin{tabular}{p{12.5cm}}
    \toprule
    You are a helpful medical coder and expert in the International Classification of Primary Care. You will receive a query and a list of results from an ICPC search engine. Your  task is to select the result that best matches the query. Your response should be a single ICPC code between the XML tags <answer>selected\_code</answer>. If there is no result good enough to match the given query, return an empty answer: <answer></answer>.\\
    \\
    Query: \{query\}\\
    \\
    Search engine results: \{search\_engine\_results\}\\
    \bottomrule
    \end{tabular}
    \label{tab:prompt}
\end{table}

\subsection{Evaluation metrics}

Several pieces of information were gathered based on the models' responses. 

The F1-score \cite{f1scorewiki} was used as the main metric to assess model performance. It is defined as:

\begin{align}
    \label{eq:f1score}
    F1 = \frac{2TP}{2TP + FP + FN}
\end{align}

with $TP$ as the number of relevant codes correctly selected by the model (true positive); $FP$ as the number of incorrectly selected codes (false positive); and $FN$ referring to cases where the model fails to provide an answer, even though a relevant code was present in the search results (false negative). The F1-score was computed for each model and for each top $k$ result from the search engine. 

To simulate an ideal retriever, F1-score was also computed considering only the cases in which there was a relevant result among the search results. The results were compared to the original performance to estimate how much it can be improved by only optimizing the search engine.

Each model received each query with a list of varying length containing the search results. Their performance was evaluated given a list of the top 10, 20, 50, 100, and, for some models, 200 results. Cost was a limiting factor for extending further the size of the search results, and only some of the top-performing models were also evaluated using the 200 top results in the prompt. 

Other aspects of model response were analyzed, including precision, recall, abstention rate \textemdash model opts to not select any code \textemdash, token usage, time for answer generation, proportion of completions within the required format, proportion of valid ICPC-2 codes, and the proportion of selected ICPC-2 codes that were present in the search engine results. 

As a reference, a baseline F1-score was computed by selecting the first result of the search engine.

The choice of the F1-score as the primary evaluation metric was deliberate, intended to reflect the nature of the task as a single operational point decision problem. The F1-score, defined in equation \ref{eq:f1score} as the harmonic mean of precision and recall, assesses the model's performance at a specific, implicit decision threshold, which corresponds to its final choice of a single code (or none).

Although the Area Under the Receiver Operating Characteristic Curve (AUC) is a widely used metric for evaluating the performance of classification models \cite{Bowers2019-su}, its interpretation can be less straightforward in practical contexts that require a single, unambiguous decision. While valuable for comparing the general ranking power of different models, it does not report on the performance at a specific operational point, which is what determines the final utility in many practical applications \cite{Tharwat2021-qo}.

As the task in this study is to assess the quality of the model's final and singular recommendation, the F1-score becomes a more direct and representative measure of performance \cite{powers2020evaluationprecisionrecallfmeasure}. Also, the absence of confidence scores in models' predictions, especially those hosted in cloud services, prohibits calibration procedures.

\subsection{Statistical analysis}

To assess the strength and direction of the monotonic association between the model performance variables, the Spearman's rank correlation test was employed. This is a non-parametric statistical method that measures the relationship between two variables by using their ranks instead of their raw values. By not assuming a specific data distribution, the Spearman test is robust to outliers and ideal for identifying relationships that are consistently increasing or decreasing, though not necessarily linear \cite{Hollander2013-wf}.

%

%
%
%
%

The analysis was performed using the implementation available in the SciPy package for Python \cite{scipySpearmanrx2014}, which properly handles ties in the data. A significance level (alpha) of 0.05 was adopted to reject the null hypothesis of no correlation. The findings were supplemented by graphical visualizations to aid interpretation.

\section{Results}

The baseline F1-score (automatically selecting the first search result) was 0.8044. The performance of the LLMs in selecting relevant ICPC-2 codes was measured using the F1-score (see Tables \ref{tab:f1score} and \ref{tab:f1scoreideal}). Table \ref{tab:f1score} shows the scores for all cases; Table \ref{tab:f1scoreideal} covers cases where the search engine actually retrieved a relevant code (ideal retriever). The three best-performing models were gpt-4.5-preview, o3, and gemini-2.5-pro-exp-03-25. In the ideal retriever context, the ranking was: o3, followed by gpt-4.5-preview and gemini-2.5-pro-exp-03-25. Detailed precision and recall metrics for each model are available in the appendix \ref{appendix}.

\begin{figure}
\caption{Relationship between mean price in USD per 1,000 responses and F1-score. For each model, the max F1-score was considered. Locally tested models were not included. Note the x-axis in log scale.}
\centering
\includegraphics[scale=0.45]{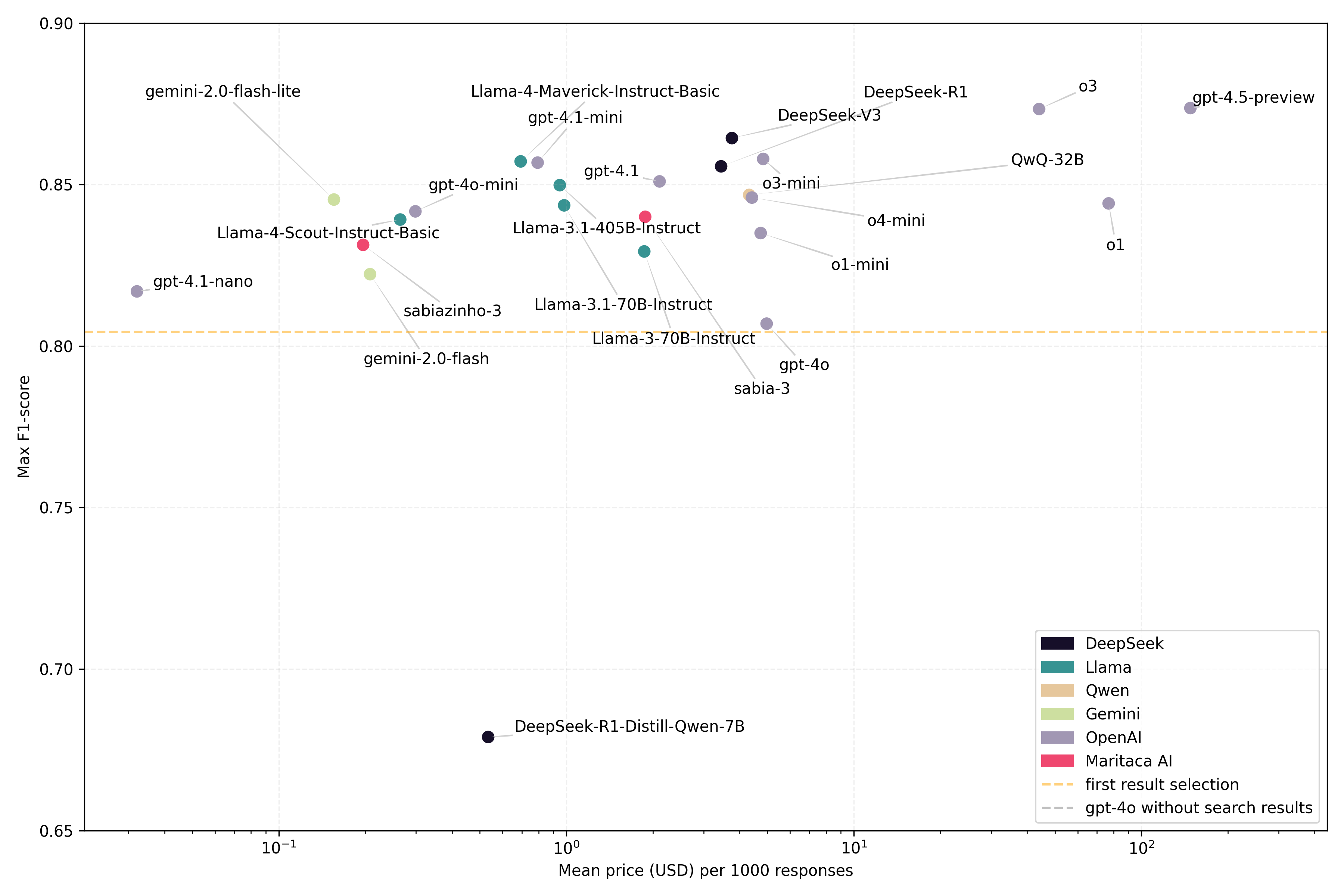}
\label{fig:pricevsf1}
\end{figure}

Additional analyses show F1-score versus average cost (figure \ref{fig:pricevsf1}), and F1-score versus average token usage (figure \ref{fig:tokenusagevsf1}). Other plots are available in the appendix \ref{appendix}:  relative performance improvement per model with an ideal retriever (figure \ref{fig:diffpermodel}), F1-score versus mean time per response (figure \ref{fig:timevsf1}), format compliance (figure \ref{fig:rightformatanswer}), code validity (figure \ref{fig:validicpcgraph}), and F1-score versus model size (figure \ref{fig:nparamsf1}).

Correlation analysis revealed significant associations between model performance and model size, token usage, cost, and response time, especially for models up to 30B parameters. Further details were also included in the appendix \ref{appendix} for conciseness.


\vfill
\vfill

\begin{figure}
\caption{Relationship between mean token usage per response and F1-score. For each model, the max F1-score was considered.}
\centering
\includegraphics[scale=0.45]{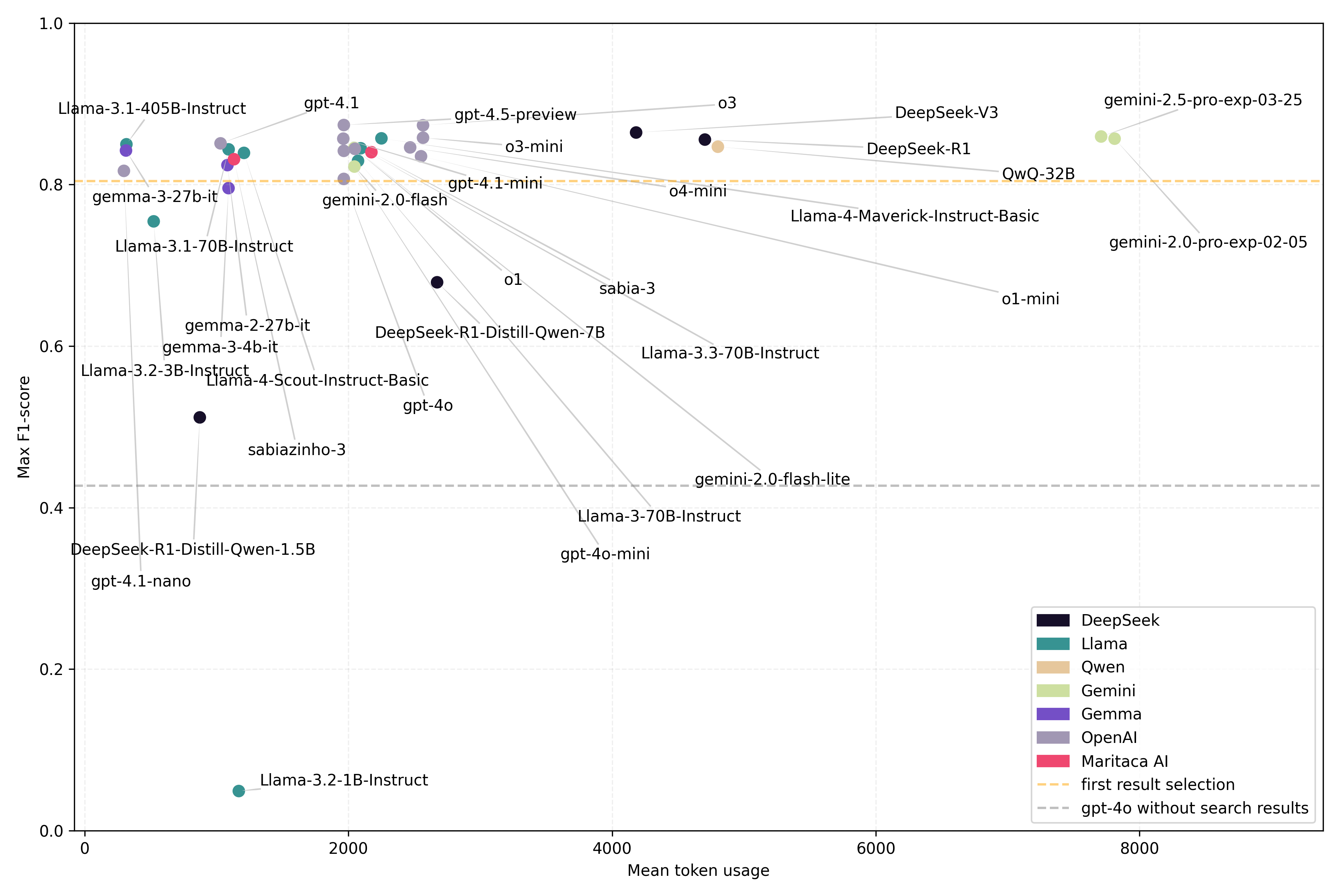}
\label{fig:tokenusagevsf1}
\end{figure}

\section{Discussion}

\subsection{Large language models as code selectors}

Several models presented good performance in the task. Out of 33 different models evaluated, 28 (85\%) obtained a max F1-score above 0.8 and 10 (30\%) obtained a max F1-score above 0.85. Among the 10 best performing models, only three are open-source: DeepSeek-V3, LLama-4-Maverick-Instruct-Basic, and DeepSeek-R1.

One of the problems of using large language models in medical coding is code hallucination or using valid codes in the wrong contexts. Lee and Lindsey \cite{lee2024largelanguagemodelsabstract} demonstrated that the representation of medical codes in large language models' knowledge is rudimentary and does not allow distinction between fake and valid codes. This study shows that, by including relevant results from an appropriate medical coding search engine, the risk of code hallucination tends to zero in most models, as shown in figure \ref{fig:validicpcgraph}. Although invalid codes were not a major issue in this study, the poor representation of medical codes described by Lee and Lindsey may be part of the problem of overconfidence in code selection. 

Overconfidence in code selection can be inferred due to the low frequency of no code selection. This is also represented by the recall being greater than precision in most models (see tables \ref{tab:f1scoredetailed} and \ref{tab:f1scoreidealdetailed}), and by examples of answers in which the model selects a code even for an unintelligible expression (see table \ref{tab:deepseekexample002}).

Most models were also able to select only valid ICPC-2 codes that were present in the list of results, showing strong capabilities for instruction following in this scenario (see figure \ref{fig:selectedcodeinresults}). These results are compatible with previous experiments with the ICD-10 classification \cite{kwan2024largelanguagemodelsgood}.

\subsection{The impact of the retriever}

When considering only the cases in which the search results include a relevant result, it is possible to estimate how much the performance can be improved by only optimizing the search engine. In this scenario, out of 33 models evaluated, 29 (88\%) obtained a max F1-score above 0.8 and 24 (73\%) obtained a max F1-score above 0.85. Among the 10 best performing models, only three are open-source: gemma-3-27b-it, DeepSeek-V3, and LLama-4-Maverick-Instruct-Basic.

As expected, the great majority of the models improve when simulating an ideal retriever, but some benefit more than others. As shown in the figure \ref{fig:diffpermodel}, the maximum observed improvement was about 4\%. The three best-performing models improved from 2\% (gpt-4.5-preview) to 4\% (gemini-2.5-pro-exp-03-25). Such improvements reflect how much the absence of a relevant result among the search results impacts the model's behavior. Most models presented higher recall and lower precision, as shown in the tables \ref{tab:f1scoredetailed} and \ref{tab:f1scoreidealdetailed}. They struggled to not select any code when no good option was available, demonstrating overconfidence in their predictions. 

\subsection{Challenges with answer formatting}

Smaller models, in particular the model Llama-3.2-1B-Instruct, had poor performance mainly due to trouble in following the requested answer format. As shown in figure \ref{fig:rightformatanswer}, this model answered in the right format only in 1.5\% of the responses. Models bigger than 4 billion parameters were more consistent in following the formatting instructions and achieved a minimum of 93.9\% of the responses with the right format. Some examples of the wrongly formatted answers of this model are available in the appendix \ref{appendix} as tables \ref{tab:llama1example001}, \ref{tab:llama1example002} and \ref{tab:llama1example003}.

\subsection{The impact of the length of results' list}

Smaller models also had their performance heavily impacted by the length of the results list. Particularly, the DeepSeek-R1-Distill-Qwen-1.5B model had a consistent decrease in performance as the results list got longer, even though it was still inside the model's context window. Bigger models had much less variability in F1-score between different list lengths, but their context window can process several thousand tokens and was not tested to the limit in this study. 

\subsection{How model performance scales}

The relationship between performance and the number of generated tokens represents the efficiency of the model in answer generation. In figure \ref{fig:tokenusagevsf1}, the closer a model is to the top-left quadrant of the plot, the better, representing a smaller quantity of tokens and a higher performance in comparison to others. Reasoning models, including o3 and DeepSeek-R1, generated 3 to 4 times the number of tokens per response with only a small relative improvement in performance. Possibly, the medical coding task with ICPC-2 was not present in the training of these reasoning models. Therefore, their reasoning may not be optimized for the correct code selection and may generate noise.

Some models with very different sizes had very close performance, suggesting that, in some cases, model size may have a weak correlation with performance. The figure \ref{fig:nparamsf1} shows that, among open-source models, there is a strong correlation between model performance and model size until the size of 30 billion parameters. Then, that correlation gets very weak. As an example, considering models with an F1-score of at least 0.8 and at most 1,000 tokens per response on average, we have Llama-3.1-405B-Instruct, with 405 billion parameters; gemma-3-27b-it, with 27 billion parameters and 15 times smaller than the first; and gpt-4.1-nano, with an unknown size. 

\subsection{Insights from reasoning tokens}

Additional insights can be drawn from inspecting the generated responses. DeepSeek-R1 responses, in particular, revealed interesting model behaviors, including: considering every code present in the search results, interpreting the query in different ways to find similar concepts, and translating the query from English to Brazilian Portuguese or the other way around. 

Some DeepSeek-R1’s responses revealed imprecisions in the evaluation dataset. This was particularly evident in situations where the vector proximity in the embedding space  caused multiple codes with similar meanings to be retrieved for a single ambiguous query. Instead of treating these terms as simple synonyms, the model applied semantic reasoning to analyze the nuances between the candidates. For instance, as seen in Table \ref{tab:deepseekexample001}, for the query ‘uso de droga’ (drug use), the model was presented with codes for both recreational drug usage and substance abuse disorder. Although only one was considered relevant in the evaluation dataset, the model's explicit reasoning process led it to consider both options, thereby revealing a mistake in the data annotation. This behavior reveals a potential application of reasoning models in large-scale data annotation for spotting inconsistencies in human annotation. Additionally, other responses revealed imprecisions in the Brazilian Portuguese thesaurus (see \ref{tab:deepseekexample003}) and how reasoning models can easily get lost with less meaningful queries (see \ref{tab:deepseekexample002}).

\subsection{Strenghts and limitations}

This study has several strengths. To date, it is the first to evaluate the ability of language models to automatically select an ICPC-2 code using the results of a semantic search. It is also the first study focused on performing this task in Brazilian Portuguese. The use of search expressions submitted by real users increases the representativeness of the findings, and the use of the official Brazilian Portuguese ICPC-2 thesaurus ensures technical rigor in the mapping of expressions to ICPC-2 codes. The findings also highlight some of the strengths and limitations of language models in automatic code selection, and indicate areas where optimization may be possible—whether in the mapping of expressions to codes or in the annotation of the data used for evaluation. The presence of models achieving F1-scores above 0.8 suggests that automatic coding of clinical records is a possibility in the near future.

This perspective is aligned with recent studies exploring LLMs for zero-shot coding of EHRs. Chen \textit{et al.} \cite{chen2024zeroshotatccodinglarge} found that while models like GPT are not yet ready for fully autonomous use due to clinically significant errors, they demonstrate strong potential as an assistive tool to augment the work of human coders.

This study also has several limitations. The number of search results included in the prompt had a greater impact on the performance of models with fewer than 3 billion parameters. However, the context window tested was much smaller than the maximum supported by many of the models. Prompts with 200 search results reached up to approximately 4,000 input tokens, and including the response brought the total number of processed tokens to around 5,000. Some of the evaluated models are capable of processing up to 10 million tokens. Including much longer lists of results was not feasible in this study due to increased cost. However, this cost is also an important limiting factor for deploying systems that process long texts in healthcare services. Therefore, models that can achieve the best performance with the most efficient use of context are essential for the practical utility of this technology.

The evaluation dataset contains a small sample of just 437 expressions annotated with a list of relevant ICPC-2 codes, and some codes are not represented, as shown in Figures \ref{fig:heatmap_2} and \ref{fig:heatmap_1}. Expanding this dataset in future studies would allow for a more detailed evaluation of model performance. Despite the limited number, the figures show that codes from all ICPC-2 chapters are represented.

From the analysis of model responses, errors and inaccuracies were found both in the mappings of the Brazilian thesaurus and in the selection of relevant codes in the evaluation dataset. A full review of all mappings would require a lot of time and effort. The model-generated responses in this study, particularly from the DeepSeek-R1 model, may help identify inconsistencies and guide necessary corrections. These errors may have influenced the results presented here and should be addressed in future work.

This study did not assess the impact of a code’s position in the result list on model performance. It is known that language models exhibit positional bias in multiple-choice tasks. Zhen et al. \cite{zheng2024largelanguagemodelsrobust} described this bias in the context of questions where options are labeled as choices A, B, etc. According to the authors, this bias arises from the probabilities associated with the tokens that label the choices, rather than the content itself. In our study, the result list was inserted into the prompt in JSON format, a ubiquitous data structure on the internet, and the response was requested by mentioning the code itself. Nevertheless, we cannot exclude the possibility that models may favor some codes over others regardless of relevance to the query. Investigating this bias was beyond the scope of this work.

It was not possible to precisely evaluate the relationship between performance and response time, as inference was conducted on different cloud platforms and on local devices. A fair comparison would require all models to be run on identical hardware. Additionally, some models accessed via API services may impose request limits, throttle token generation speeds, or use caching mechanisms to accelerate responses. Response time may also vary depending on the hardware used and internet connection speed.

\subsection{Future directions}

Addressing large language models' positional bias \cite{zheng2024largelanguagemodelsrobust} in code selection was beyond the scope of this study. Future studies may explore how this limitation impacts medical coding by randomizing items' order in the JSON list and observing its impact on performance.

Only a single simple prompt was used to evaluate all of the models. Addressing prompting strategies (few-shot examples, chain-of-thought prompting) and optimizing the prompt for each model can result in performance gains and may be contemplated in future work. 

According to this study, these models struggle to return an empty answer even when there is no relevant code to select. Future studies should explore different techniques to address this problem, such as different prompting techniques and fine-tuning with positive and negative examples. 

Model distillation is an alternative since it compresses a bigger model's knowledge into a smaller "student" model. Instead of only training the smaller model on the "correct" final codes (a process known as fine-tuning), distillation trains the student model to replicate the entire probability distribution of the larger "teacher" model. Consequently, a distilled smaller model could potentially learn to better adhere to the required output format and avoid generating codes not present in the provided context, effectively addressing the deficiencies observed in models with fewer than 3 billion parameters. 

Finally, medical coding can be framed as a task with verifiable answers and may benefit from fine-tuning strategies involving reinforcement learning with verifiable rewards, such as GRPO \cite{shao2024deepseekmathpushinglimitsmathematical}, DAPO \cite{yu2025dapoopensourcellmreinforcement} and VAPO \cite{yue2025vapoefficientreliablereinforcement}. 

\section{Conclusions}

This study demonstrates that large language models (LLMs) can effectively automate the selection of ICPC-2 codes from clinical expressions using semantic search outputs. Many models—particularly large proprietary ones—achieved high F1-scores above 0.8, although model size did not consistently correlate with efficiency. Notably, smaller models (with fewer than 3 billion parameters) struggled with formatting adherence and hallucination control.

Most models successfully followed formatting instructions and avoided generating invalid codes. Optimizing the semantic search engine alone showed potential to improve code selection performance up to 4\%. Model efficiency, in terms of token usage, cost, and latency, varied substantially and should be considered in practical deployments. Analysis of DeepSeek-R1's responses uncovered potential inconsistencies in the dataset, suggesting LLMs may assist in improving data annotation quality.

These findings provide a performance baseline for the development of future ICPC-2-specific coding models. Further work may explore domain-specific training, improved prompting techniques, and fine-tuning strategies to enhance performance, reduce costs, and enable broader applicability.

\begin{ack}

\subsection*{Author contributions}

Vinicius Almeida: conceived and designed the study; collected the data; contributed data and analysis tools; performed the analysis; wrote the paper; wrote the source code. Vinicius de Camargo: contributed to paper review and critical feedback. Raquel Gómez-Bravo: contributed to paper review and critical feedback. Egbert van der Haring: contributed to paper review and critical feedback. Kees van Boven: contributed to paper review and critical feedback. Marcelo Finger: contributed to paper review and critical feedback. Luis Lopez: contributed to paper review and critical feedback. 

\subsection*{Ethical statement}
The study was approved by the institutional ethics committee of Hospital das Clínicas da Faculdade de Medicina da Universidade de São Paulo (HCFMUSP) (No. 70023923.3.0000.0068).

\subsection*{Conflicts of Interest}
None.

\subsection*{Funding}
None.

\end{ack}

\bibliographystyle{unsrt}
\bibliography{references}

\newpage
\appendix
\renewcommand{\thefigure}{A.\arabic{figure}} 
\renewcommand{\thetable}{A.\arabic{table}} 
\setcounter{figure}{0} 
\setcounter{table}{0} 

\section{Appendix / supplemental material}
\label{appendix}

\subsection{Hardware}
\label{section:hardware}

This study was conducted on a desktop equipped with a 13th Gen Intel(R) Core(TM) i9-13900K (3.00 GHz, 24 cores), 64 GB DDR5 RAM, and a NVIDIA GeForce RTX 4090 GPU with 24 GB VRAM. The system had a 4 TB Kingston SSD for storage and Windows 11 Pro as the operating system.

\subsection{Source code and data availability}

The source code and all data used in this study are available through a public GitHub repository at \href{https://github.com/almeidava93/llm-as-code-selectors-paper}{https://github.com/almeidava93/llm-as-code-selectors-paper} \cite{Anjos_de_Almeida_Source_code_for_2025}.

\subsection{Additional plots and tables}
This appendix includes additional plots and tables to give a more comprehensive understanding of this study's methodology. 

\begin{landscape}
\begin{figure}
\centering
\includegraphics[scale=0.45]{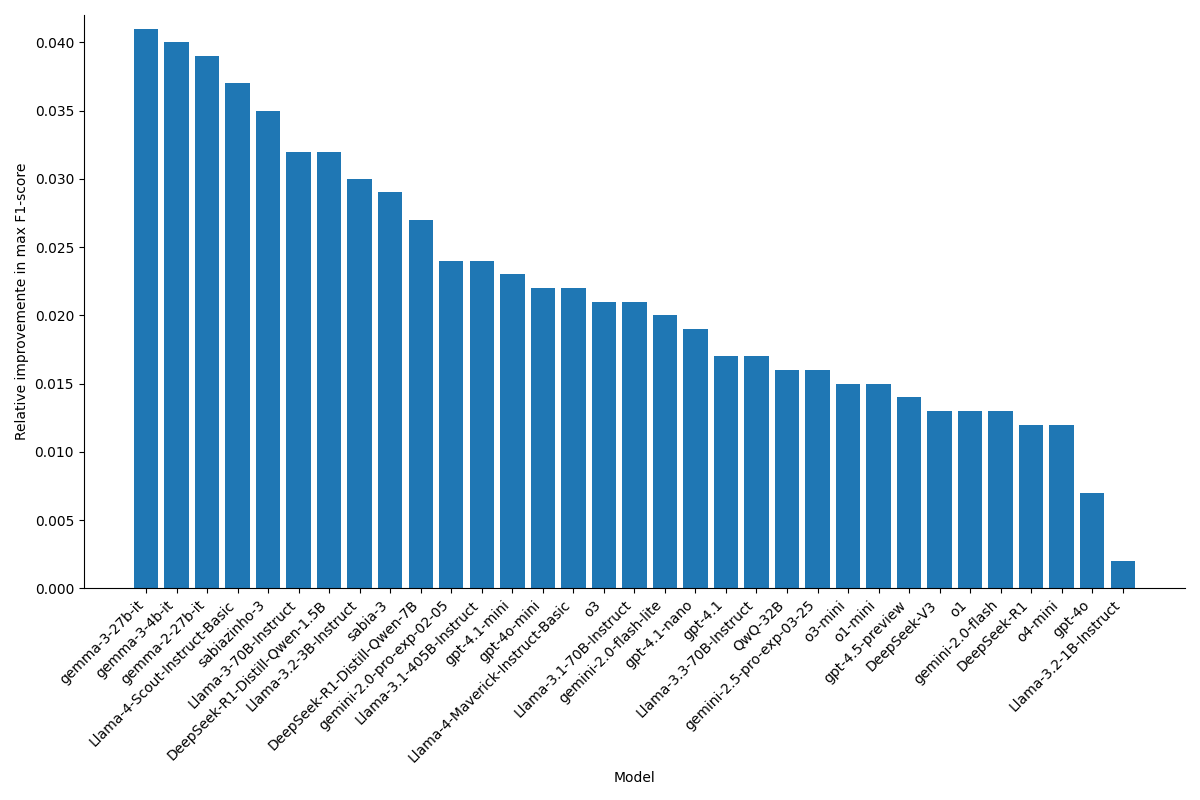}
\caption{Relative improvement in F1-score only considering cases in which there was a relevant code among the search results. For each model, the max F1-score was considered.}
\label{fig:diffpermodel}
\end{figure}
\end{landscape}

Figures \ref{fig:heatmap_1} and \ref{fig:heatmap_2} contain a frequency map of the ICPC-2 codes represented in the evaluation dataset. Figure \ref{fig:diffpermodel} shows the estimated relative improvement in performance with an ideal retriever. Figure \ref{fig:timevsf1} shows the relationship between performance and mean time per response for each model. Figure \ref{fig:rightformatanswer} shows the proportion of answers that followed the required format for each model. Figure \ref{fig:validicpcgraph} shows the proportion of answers that had valid ICPC-2 codes. Figure \ref{fig:selectedcodeinresults} shows the proportion of answers that had valid ICPC-2 codes and these were present in the search engine results. Figure \ref{fig:nparamsf1} shows the relationship between model size and performance.

Tables \ref{tab:listofllmsopen} and \ref{tab:listofllmsclosed} list all the evaluated models, their price, size if available, and hosting platform. Tables \ref{tab:f1score} and \ref{tab:f1scoreideal} contain detailed metrics related to each model's performance, both with the standard retriever and the ideal retriever simulation. Table \ref{tab:promptexample} includes a complete example of a prompt used to interact with each large language model. Tables \ref{tab:f1scoredetailed} and \ref{tab:f1scoreidealdetailed} show details about precision, recall, and F1-score for each model and each top $k$ results from the search engine. Tables \ref{tab:llama1example001}, \ref{tab:llama1example002} and \ref{tab:llama1example003} show different examples mentioned in the article of responses from the Llama-3.2-1B-Instruct model. Table \ref{tab:correlation} contains the results of the Spearman correlation test. Tables \ref{tab:deepseekexample001}, \ref{tab:deepseekexample003} and \ref{tab:deepseekexample002} contain the examples of DeepSeek-R1's responses mentioned in the article.

\newpage

\begin{table}[H]
\centering
\begin{threeparttable}
\caption{Open-source large language models evaluated.}%
\begin{tabular}{llcc}%
\toprule%
Provider&Model name&N. parameters\tnote{1}&Price (input/output)\tnote{2}\\%
\midrule%
\addlinespace[10pt]%
Meta&Llama-4-Scout-Instruct-Basic&109 B&0.15 / 0.60\tnote{4}\\
&Llama-4-Maverick-Instruct-Basic&400 B&0.22 / 0.88\tnote{4}\\
&Llama-3.3-70B-Instruct&70 B&- \tnote{5}\\
&Llama-3.2-3B-Instruct&3 B&- \tnote{3}\\
&Llama-3.2-1B-Instruct&1 B&- \tnote{3}\\
&Llama-3.1-405B-Instruct&405 B& 3.00 / 3.00\tnote{4}\\
&Llama-3.1-70B-Instruct&70 B& 0.90 / 0.90\tnote{4}\\
&Llama-3-70B-Instruct&70 B& 0.90 / 0.90\tnote{4}\\
\addlinespace[10pt]%
Google&gemma-3-27b-it&27 B&- \tnote{5}\\
&gemma-3-4b-it&4 B&- \tnote{3}\\
&gemma-2-27b-it&27 B&- \tnote{5}\\
\addlinespace[10pt]%
Qwen&QwQ-32B&32 B&0.90 / 0.90\tnote{4}\\
\addlinespace[10pt]%
DeepSeek&DeepSeek-R1&685 B&0.55 / 2.19\tnote{4}\\
&DeepSeek-V3&685 B&0.9 / 0.9\tnote{4}\\
&DeepSeek-R1-Distill-Qwen-7B&7 B&0.20 / 0.20\tnote{4}\\
&DeepSeek-R1-Distill-Qwen-1.5B&1.5 B&- \tnote{3}\\
\bottomrule
\end{tabular}%
\begin{tablenotes}
   \item [1] Billions of parameters.
   \item [2] US dollars per million tokens (USD/M). BRL to USD conversion rate set to 1 USD = 5.8 BRL.
   \item [3] Local inference.
   \item [4] Price based on the hosting platform \textit{Fireworks.ai} \cite{fireworksFireworksFastest}.
   \item [5] Price based on the hosting platform \textit{HuggingFace} \cite{huggingfaceHuggingFace}. In this study, it was not possible to estimate.
 \end{tablenotes}
\label{tab:listofllmsopen}%
\end{threeparttable}
\end{table}

\begin{table}[H]
\centering
\begin{threeparttable}
\caption{Private large language models evaluated.}%
\begin{tabular}{llcc}%
\toprule%
Provider&Model name&N. parameters\tnote{1}&Price (input/output)\tnote{2}\\%
\midrule%
\addlinespace[10pt]%
OpenAI&gpt-4o-mini&- \tnote{3}&0.15 / 0.60\\
&gpt-4o&-&2.50 / 10.00\\
&gpt-4.5-preview&-&75.00 / 150.00\\
&o1-mini&-&1.10 / 4.40\\
&o1&-&15.00 / 60.00\\
&o3&-&10.00 / 40.00\\
&o3-mini&-&1.10 / 4.40\\
&gpt-4.1&-&2.00 / 8.00\\
&gpt-4.1-mini&-&0.40 / 1.60\\
&gpt-4.1-nano&-&0.10 / 0.40\\
&o4-mini&-&1.10 / 4.40\\
\addlinespace[10pt]%
Google&gemini-2.0-flash&-&0.10 / 0.40\\
&gemini-2.0-flash-lite&-&0.075 / 0.30\\
&gemini-2.0-pro-exp-02-05&-&- \tnote{4}\\
&gemini-2.5-pro-exp-03-25&-&- \tnote{4}\\
\addlinespace[10pt]%
Maritaca AI&sabia-3&-&0.86 / 1.72\\
&sabiazinho-3&-&0.17 / 0.52\\
\addlinespace[10pt]%
\bottomrule
\end{tabular}%
\begin{tablenotes}
   \item [1] Billions of parameters.
   \item [2] US dollars per million tokens (USD/M). BRL to USD conversion rate set to 1 USD = 5.8 BRL.
   \item [3] Number of parameters not disclosed.
   \item [4] Experimental model available free of charge at the moment of this study.
\end{tablenotes}
\label{tab:listofllmsclosed}%
\end{threeparttable}
\end{table}

\begin{longtable}{p{3.8cm}ccccccc}
\caption{F1-score of each model with each top $k$ search results list. Mean and max values are also provided. The results are sorted by the max F1-score value in descending order. The highest score of each column is in bold.} 
\label{tab:f1score}
\\
\toprule
Model / top $k$ & 10 & 20 & 50 & 100 & 200 & Mean & Max \\\midrule
\endfirsthead
\toprule
Model / top $k$ & 10 & 20 & 50 & 100 & 200 & Mean & Max \\\midrule
\endhead
\midrule
\multicolumn{8}{r}{Continues in the next page} \\
\midrule
\endfoot
\bottomrule
\endlastfoot
gpt-4.5-preview & 0.8564 & \textbf{0.8595} & \textbf{0.8671} & \textbf{0.8737} & \textbf{0.8710} & \textbf{0.8655} & \textbf{0.8737} \\
o3 & \textbf{0.8693} & 0.8568 & 0.8525 & 0.8734 &  & 0.8630 & 0.8734 \\
gemini-2.5-pro-exp-03-25 & 0.8450 & 0.8556 & 0.8629 & 0.8678 & 0.8686 & 0.8600 & 0.8686 \\
DeepSeek-V3 & 0.8477 & 0.8501 & 0.8587 & 0.8514 & 0.8644 & 0.8545 & 0.8644 \\
gemini-2.0-pro-exp-02-05 & 0.8486 & 0.8470 & 0.8564 & 0.8622 & 0.8525 & 0.8533 & 0.8622 \\
o3-mini & 0.8501 & 0.8450 & 0.8463 & 0.8579 & 0.8533 & 0.8505 & 0.8579 \\
Llama-4-Maverick-Instruct-Basic & 0.8309 & 0.8407 & 0.8453 & 0.8571 &  & 0.8435 & 0.8571 \\
gpt-4.1-mini & 0.8506 & 0.8541 & 0.8491 & 0.8568 &  & 0.8526 & 0.8568 \\
DeepSeek-R1 & 0.8336 & 0.8395 & 0.8420 & 0.8420 & 0.8556 & 0.8425 & 0.8556 \\
gpt-4.1 & 0.8482 & 0.8408 & 0.8509 & 0.8494 &  & 0.8473 & 0.8509 \\
Llama-3.1-405B-Instruct & 0.8498 & 0.8437 & 0.8455 & 0.8486 &  & 0.8469 & 0.8498 \\
QwQ-32B & 0.8122 & 0.8194 & 0.8354 & 0.8315 & 0.8468 & 0.8291 & 0.8468 \\
o4-mini & 0.8359 & 0.8372 & 0.8347 & 0.8459 &  & 0.8384 & 0.8459 \\
gemini-2.0-flash-lite & 0.8324 & 0.8453 & 0.8438 & 0.8391 &  & 0.8402 & 0.8453 \\
Llama-3.3-70B-Instruct & 0.8377 & 0.8408 & 0.8441 & 0.8449 &  & 0.8419 & 0.8449 \\
o1 & 0.8384 & 0.8285 & 0.8441 & 0.8398 &  & 0.8377 & 0.8441 \\
Llama-3.1-70B-Instruct & 0.8301 & 0.8349 & 0.8435 & 0.8381 &  & 0.8367 & 0.8435 \\
sabia-3 & 0.8243 & 0.8306 & 0.8369 & 0.8422 &  & 0.8335 & 0.8422 \\
gemma-3-27b-it & 0.8421 & 0.8333 & 0.8322 & 0.8133 &  & 0.8302 & 0.8421 \\
gpt-4o-mini & 0.8399 & 0.8408 & 0.8397 & 0.8417 &  & 0.8405 & 0.8417 \\
Llama-4-Scout-Instruct-Basic & 0.8345 & 0.8383 & 0.8391 & 0.8360 &  & 0.8370 & 0.8391 \\
sabiazinho-3 & 0.8272 & 0.8197 & 0.8356 & 0.8324 &  & 0.8287 & 0.8356 \\
o1-mini & 0.8280 & 0.8169 & 0.8338 & 0.8350 &  & 0.8284 & 0.8350 \\
Llama-3-70B-Instruct & 0.8171 & 0.8277 & 0.8281 & 0.8293 &  & 0.8255 & 0.8293 \\
gemma-2-27b-it & 0.8072 & 0.8224 & 0.8240 & 0.8229 &  & 0.8191 & 0.8240 \\
gemini-2.0-flash & 0.8045 & 0.8158 & 0.7926 & 0.8227 &  & 0.8089 & 0.8227 \\
gpt-4.1-nano & 0.8169 & 0.7908 & 0.7989 & 0.7825 &  & 0.7972 & 0.8169 \\
gpt-4o & 0.7651 & 0.7595 & 0.7988 & 0.8069 &  & 0.7826 & 0.8069 \\
gemma-3-4b-it & 0.7938 & 0.7939 & 0.7955 & 0.7842 &  & 0.7919 & 0.7955 \\
Llama-3.2-3B-Instruct & 0.7316 & 0.7544 & 0.6453 & 0.7237 &  & 0.7137 & 0.7544 \\
DeepSeek-R1-Distill-Qwen-7B & 0.6455 & 0.5925 & 0.6728 & 0.6790 &  & 0.6474 & 0.6790 \\
DeepSeek-R1-Distill-Qwen-1.5B & 0.5117 & 0.4630 & 0.3475 & 0.1308 &  & 0.3632 & 0.5117 \\
Llama-3.2-1B-Instruct & 0.0050 & 0.0000 & 0.0491 & 0.0389 &  & 0.0233 & 0.0491 \\
\end{longtable}

\begin{longtable}{p{3.8cm}ccccccc}
\captionsetup{width=\linewidth}
\caption{F1-score of each model with each top $k$ search results list considering only the cases in which there is a relevant code among the search results. Mean and max values are also provided. The results are sorted by the max F1-score value in descending order. The highest score of each column is in bold.} 
\label{tab:f1scoreideal}
\\
\toprule
Model / top $k$ & 10 & 20 & 50 & 100 & 200 & Mean & Max \\
\midrule
\endfirsthead
\toprule
Model / top $k$ & 10 & 20 & 50 & 100 & 200 & Mean & Max \\
\midrule
\endhead
\midrule
\multicolumn{8}{r}{Continues in the next page} \\
\midrule
\endfoot
\bottomrule
\endlastfoot
o3 & \textbf{0.8919} & \textbf{0.8779} & 0.8712 & \textbf{0.8874} &  & \textbf{0.8821} & \textbf{0.8919} \\
gpt-4.5-preview & 0.8741 & 0.8748 & \textbf{0.8813} & 0.8856 & \textbf{0.8804} & 0.8793 & 0.8856 \\
gemini-2.5-pro-exp-03-25 & 0.8627 & 0.8686 & 0.8759 & 0.8820 & 0.8780 & 0.8734 & 0.8820 \\
gpt-4.1-mini & 0.8766 & 0.8764 & 0.8689 & 0.8718 &  & 0.8734 & 0.8766 \\
gemma-3-27b-it & 0.8764 & 0.8604 & 0.8579 & 0.8352 &  & 0.8575 & 0.8764 \\
DeepSeek-V3 & 0.8643 & 0.8631 & 0.8716 & 0.8618 & 0.8760 & 0.8674 & 0.8760 \\
Llama-4-Maverick-Instruct-Basic & 0.8599 & 0.8626 & 0.8638 & 0.8757 &  & 0.8655 & 0.8757 \\
gemini-2.0-flash-lite & 0.8688 & 0.8745 & 0.8681 & 0.8587 &  & 0.8675 & 0.8745 \\
gemini-2.0-pro-exp-02-05 & 0.8736 & 0.8659 & 0.8741 & 0.8728 & 0.8618 & 0.8696 & 0.8741 \\
o3-mini & 0.8679 & 0.8591 & 0.8591 & 0.8709 & 0.8626 & 0.8639 & 0.8709 \\
Llama-4-Scout-Instruct-Basic & 0.8671 & 0.8699 & 0.8694 & 0.8591 &  & 0.8664 & 0.8699 \\
Llama-3.1-405B-Instruct & 0.8698 & 0.8634 & 0.8607 & 0.8615 &  & 0.8638 & 0.8698 \\
sabia-3 & 0.8616 & 0.8607 & 0.8646 & 0.8666 &  & 0.8634 & 0.8666 \\
DeepSeek-R1 & 0.8499 & 0.8536 & 0.8571 & 0.8536 & 0.8660 & 0.8561 & 0.8660 \\
gpt-4.1 & 0.8658 & 0.8571 & 0.8650 & 0.8599 &  & 0.8620 & 0.8658 \\
sabiazinho-3 & 0.8575 & 0.8475 & 0.8611 & 0.8520 &  & 0.8545 & 0.8611 \\
Llama-3.1-70B-Instruct & 0.8596 & 0.8596 & 0.8611 & 0.8532 &  & 0.8584 & 0.8611 \\
QwQ-32B & 0.8305 & 0.8357 & 0.8492 & 0.8476 & 0.8607 & 0.8447 & 0.8607 \\
gpt-4o-mini & 0.8599 & 0.8571 & 0.8560 & 0.8544 &  & 0.8569 & 0.8599 \\
Llama-3.3-70B-Instruct & 0.8552 & 0.8583 & 0.8596 & 0.8543 &  & 0.8568 & 0.8596 \\
gemma-2-27b-it & 0.8410 & 0.8563 & 0.8507 & 0.8436 &  & 0.8479 & 0.8563 \\
o4-mini & 0.8559 & 0.8524 & 0.8476 & 0.8552 &  & 0.8528 & 0.8559 \\
Llama-3-70B-Instruct & 0.8510 & 0.8555 & 0.8536 & 0.8488 &  & 0.8522 & 0.8555 \\
o1 & 0.8539 & 0.8402 & 0.8547 & 0.8468 &  & 0.8489 & 0.8547 \\
o1-mini & 0.8469 & 0.8321 & 0.8479 & 0.8455 &  & 0.8431 & 0.8479 \\
gemini-2.0-flash & 0.8207 & 0.8274 & 0.8029 & 0.8331 &  & 0.8210 & 0.8331 \\
gpt-4.1-nano & 0.8321 & 0.8012 & 0.8105 & 0.7917 &  & 0.8089 & 0.8321 \\
gemma-3-4b-it & 0.8275 & 0.8249 & 0.8208 & 0.8023 &  & 0.8189 & 0.8275 \\
gpt-4o & 0.7744 & 0.7688 & 0.8071 & 0.8128 &  & 0.7907 & 0.8128 \\
Llama-3.2-3B-Instruct & 0.7702 & 0.7771 & 0.6579 & 0.7404 &  & 0.7364 & 0.7771 \\
DeepSeek-R1-Distill-Qwen-7B & 0.6744 & 0.6156 & 0.6930 & 0.6972 &  & 0.6701 & 0.6972 \\
DeepSeek-R1-Distill-Qwen-1.5B & 0.5279 & 0.4762 & 0.3525 & 0.1315 &  & 0.3720 & 0.5279 \\
Llama-3.2-1B-Instruct & 0.0050 & 0.0000 & 0.0493 & 0.0391 &  & 0.0234 & 0.0493 \\
\end{longtable}

\begin{table}[h]
    \centering
    \caption{Complete example of a prompt used to interact with each large language model.}
    \begin{tabular}{p{12.5cm}}
    \toprule
    You are a helpful medical coder and expert in the International Classification of Primary Care. You will receive a query and a list of results from an ICPC search engine. Your  task is to select the result that best matches the query. Your response should be a single ICPC code between the XML tags <answer>selected\_code</answer>. If there is no result good enough to match the given query, return an empty answer: <answer></answer>.\\
    \\
    Query: dor coluna\\
    \\
    Search engine results: [\{`code': `L03', `expression': `dor lombar'\}, \{`code': `L02', `expression': `dor nas costas'\}, \{`code': `L02', `expression': `dor vertebral'\}, \{`code': `L02', `expression': `dor no dorso'\}, \{`code': `L01', `expression': `dor cervical'\}, \{`code': `L02', `expression': `dores nas costas'\}, \{`code': `L03', `expression': `dor vertebral lombar baixa'\}, \{`code': `L03', `expression': `dor dorsal baixa'\}, \{`code': `L02', `expression': `dor vertebral torácica'\}, \{`code': `L86', `expression': `dor ciática'\}]\\
    \bottomrule
    \end{tabular}
    \label{tab:promptexample}
\end{table}

\begin{longtable}{lrrrr}
\captionsetup{width=\linewidth}
\caption{Detailed results including precision and recall of each model in each top $k$ results.} 
\label{tab:f1scoredetailed}
\\
\toprule
Model & top $k$ & Precision & Recall & F1-score \\
\midrule
\endfirsthead
\toprule
Model & top $k$ & Precision & Recall & F1-score \\
\midrule
\endhead
\midrule
\multicolumn{5}{r}{Continues in the next page} \\
\midrule
\endfoot
\bottomrule
\endlastfoot

DeepSeek-R1 & 10 & 0.7891 & 0.8834 & 0.8336 \\
DeepSeek-R1 & 20 & 0.7826 & 0.9053 & 0.8395 \\
DeepSeek-R1 & 50 & 0.7803 & 0.9142 & 0.8420 \\
DeepSeek-R1 & 100 & 0.7725 & 0.9251 & 0.8420 \\
DeepSeek-R1 & 200 & 0.7921 & 0.9302 & 0.8556 \\
DeepSeek-R1-Distill-Qwen-1.5B & 10 & 0.5569 & 0.4733 & 0.5117 \\
DeepSeek-R1-Distill-Qwen-1.5B & 20 & 0.5187 & 0.4181 & 0.4630 \\
DeepSeek-R1-Distill-Qwen-1.5B & 50 & 0.5658 & 0.2507 & 0.3475 \\
DeepSeek-R1-Distill-Qwen-1.5B & 100 & 0.6087 & 0.0733 & 0.1308 \\
DeepSeek-R1-Distill-Qwen-7B & 10 & 0.5126 & 0.8712 & 0.6455 \\
DeepSeek-R1-Distill-Qwen-7B & 20 & 0.4502 & 0.8660 & 0.5925 \\
DeepSeek-R1-Distill-Qwen-7B & 50 & 0.5394 & 0.8939 & 0.6728 \\
DeepSeek-R1-Distill-Qwen-7B & 100 & 0.5511 & 0.8840 & 0.6790 \\
DeepSeek-V3 & 10 & 0.7903 & 0.9142 & 0.8477 \\
DeepSeek-V3 & 20 & 0.7899 & 0.9204 & 0.8501 \\
DeepSeek-V3 & 50 & 0.7935 & 0.9355 & 0.8587 \\
DeepSeek-V3 & 100 & 0.7778 & 0.9403 & 0.8514 \\
DeepSeek-V3 & 200 & 0.7888 & 0.9559 & 0.8644 \\
Llama-3-70B-Instruct & 10 & 0.7038 & 0.9738 & 0.8171 \\
Llama-3-70B-Instruct & 20 & 0.7227 & 0.9683 & 0.8277 \\
Llama-3-70B-Instruct & 50 & 0.7234 & 0.9684 & 0.8281 \\
Llama-3-70B-Instruct & 100 & 0.7251 & 0.9684 & 0.8293 \\
Llama-3.1-405B-Instruct & 10 & 0.7792 & 0.9345 & 0.8498 \\
Llama-3.1-405B-Instruct & 20 & 0.7597 & 0.9485 & 0.8437 \\
Llama-3.1-405B-Instruct & 50 & 0.7723 & 0.9341 & 0.8455 \\
Llama-3.1-405B-Instruct & 100 & 0.7753 & 0.9373 & 0.8486 \\
Llama-3.1-70B-Instruct & 10 & 0.7337 & 0.9558 & 0.8301 \\
Llama-3.1-70B-Instruct & 20 & 0.7409 & 0.9563 & 0.8349 \\
Llama-3.1-70B-Instruct & 50 & 0.7617 & 0.9451 & 0.8435 \\
Llama-3.1-70B-Instruct & 100 & 0.7512 & 0.9477 & 0.8381 \\
Llama-3.2-1B-Instruct & 10 & 0.2500 & 0.0025 & 0.0050 \\
Llama-3.2-1B-Instruct & 20 & 0.0000 & 0.0000 & 0.0000 \\
Llama-3.2-1B-Instruct & 50 & 0.5556 & 0.0257 & 0.0491 \\
Llama-3.2-1B-Instruct & 100 & 0.2963 & 0.0208 & 0.0389 \\
Llama-3.2-3B-Instruct & 10 & 0.6154 & 0.9018 & 0.7316 \\
Llama-3.2-3B-Instruct & 20 & 0.6615 & 0.8776 & 0.7544 \\
Llama-3.2-3B-Instruct & 50 & 0.6361 & 0.6547 & 0.6453 \\
Llama-3.2-3B-Instruct & 100 & 0.6496 & 0.8169 & 0.7237 \\
Llama-3.3-70B-Instruct & 10 & 0.7714 & 0.9164 & 0.8377 \\
Llama-3.3-70B-Instruct & 20 & 0.7667 & 0.9307 & 0.8408 \\
Llama-3.3-70B-Instruct & 50 & 0.7786 & 0.9217 & 0.8441 \\
Llama-3.3-70B-Instruct & 100 & 0.7821 & 0.9187 & 0.8449 \\
Llama-4-Maverick-Instruct-Basic & 10 & 0.7292 & 0.9654 & 0.8309 \\
Llama-4-Maverick-Instruct-Basic & 20 & 0.7476 & 0.9602 & 0.8407 \\
Llama-4-Maverick-Instruct-Basic & 50 & 0.7602 & 0.9520 & 0.8453 \\
Llama-4-Maverick-Instruct-Basic & 100 & 0.7660 & 0.9730 & 0.8571 \\
Llama-4-Scout-Instruct-Basic & 10 & 0.7329 & 0.9688 & 0.8345 \\
Llama-4-Scout-Instruct-Basic & 20 & 0.7352 & 0.9749 & 0.8383 \\
Llama-4-Scout-Instruct-Basic & 50 & 0.7279 & 0.9905 & 0.8391 \\
Llama-4-Scout-Instruct-Basic & 100 & 0.7283 & 0.9811 & 0.8360 \\
QwQ-32B & 10 & 0.7481 & 0.8882 & 0.8122 \\
QwQ-32B & 20 & 0.7468 & 0.9077 & 0.8194 \\
QwQ-32B & 50 & 0.7580 & 0.9303 & 0.8354 \\
QwQ-32B & 100 & 0.7482 & 0.9358 & 0.8315 \\
QwQ-32B & 200 & 0.7536 & 0.9663 & 0.8468 \\
gemini-2.0-flash & 10 & 0.7715 & 0.8466 & 0.8073 \\
gemini-2.0-flash & 20 & 0.7737 & 0.8698 & 0.8189 \\
gemini-2.0-flash & 50 & 0.7742 & 0.8446 & 0.8079 \\
gemini-2.0-flash & 100 & 0.7769 & 0.8732 & 0.8222 \\
gemini-2.0-flash-lite & 10 & 0.7262 & 0.9621 & 0.8277 \\
gemini-2.0-flash-lite & 20 & 0.7440 & 0.9569 & 0.8371 \\
gemini-2.0-flash-lite & 50 & 0.7500 & 0.9574 & 0.8411 \\
gemini-2.0-flash-lite & 100 & 0.7512 & 0.9665 & 0.8453 \\
gemini-2.0-pro-exp-02-05 & 10 & 0.7673 & 0.9281 & 0.8401 \\
gemini-2.0-pro-exp-02-05 & 20 & 0.7567 & 0.9453 & 0.8405 \\
gemini-2.0-pro-exp-02-05 & 50 & 0.7632 & 0.9696 & 0.8541 \\
gemini-2.0-pro-exp-02-05 & 100 & 0.7724 & 0.9522 & 0.8529 \\
gemini-2.0-pro-exp-02-05 & 200 & 0.7727 & 0.9613 & 0.8568 \\
gemini-2.5-pro-exp-03-25 & 10 & 0.7679 & 0.9284 & 0.8405 \\
gemini-2.5-pro-exp-03-25 & 20 & 0.7659 & 0.9486 & 0.8475 \\
gemini-2.5-pro-exp-03-25 & 50 & 0.7614 & 0.9634 & 0.8506 \\
gemini-2.5-pro-exp-03-25 & 100 & 0.7835 & 0.9499 & 0.8587 \\
gemini-2.5-pro-exp-03-25 & 200 & 0.7751 & 0.9643 & 0.8594 \\
gemma-2-27b-it & 10 & 0.6896 & 0.9732 & 0.8072 \\
gemma-2-27b-it & 20 & 0.7049 & 0.9869 & 0.8224 \\
gemma-2-27b-it & 50 & 0.7089 & 0.9837 & 0.8240 \\
gemma-2-27b-it & 100 & 0.7106 & 0.9773 & 0.8229 \\
gemma-3-27b-it & 10 & 0.7411 & 0.9750 & 0.8421 \\
gemma-3-27b-it & 20 & 0.7332 & 0.9652 & 0.8333 \\
gemma-3-27b-it & 50 & 0.7297 & 0.9683 & 0.8322 \\
gemma-3-27b-it & 100 & 0.6950 & 0.9800 & 0.8133 \\
gemma-3-4b-it & 10 & 0.7075 & 0.9042 & 0.7938 \\
gemma-3-4b-it & 20 & 0.7020 & 0.9135 & 0.7939 \\
gemma-3-4b-it & 50 & 0.7082 & 0.9073 & 0.7955 \\
gemma-3-4b-it & 100 & 0.6864 & 0.9145 & 0.7842 \\
gpt-4.1 & 10 & 0.7924 & 0.9125 & 0.8482 \\
gpt-4.1 & 20 & 0.7764 & 0.9169 & 0.8408 \\
gpt-4.1 & 50 & 0.7850 & 0.9290 & 0.8509 \\
gpt-4.1 & 100 & 0.7864 & 0.9233 & 0.8494 \\
gpt-4.1-mini & 10 & 0.7745 & 0.9433 & 0.8506 \\
gpt-4.1-mini & 20 & 0.7687 & 0.9608 & 0.8541 \\
gpt-4.1-mini & 50 & 0.7571 & 0.9666 & 0.8491 \\
gpt-4.1-mini & 100 & 0.7709 & 0.9642 & 0.8568 \\
gpt-4.1-nano & 10 & 0.7733 & 0.8657 & 0.8169 \\
gpt-4.1-nano & 20 & 0.7569 & 0.8278 & 0.7908 \\
gpt-4.1-nano & 50 & 0.7658 & 0.8348 & 0.7989 \\
gpt-4.1-nano & 100 & 0.7679 & 0.7976 & 0.7825 \\
gpt-4.5-preview & 10 & 0.8061 & 0.9133 & 0.8564 \\
gpt-4.5-preview & 20 & 0.8030 & 0.9244 & 0.8595 \\
gpt-4.5-preview & 50 & 0.8055 & 0.9390 & 0.8671 \\
gpt-4.5-preview & 100 & 0.8145 & 0.9420 & 0.8737 \\
gpt-4.5-preview & 200 & 0.8120 & 0.9391 & 0.8710 \\
gpt-4o & 10 & 0.8167 & 0.7195 & 0.7651 \\
gpt-4o & 20 & 0.8071 & 0.7171 & 0.7595 \\
gpt-4o & 50 & 0.8059 & 0.7919 & 0.7988 \\
gpt-4o & 100 & 0.8069 & 0.8069 & 0.8069 \\
gpt-4o-mini & 10 & 0.7694 & 0.9247 & 0.8399 \\
gpt-4o-mini & 20 & 0.7667 & 0.9307 & 0.8408 \\
gpt-4o-mini & 50 & 0.7630 & 0.9335 & 0.8397 \\
gpt-4o-mini & 100 & 0.7585 & 0.9453 & 0.8417 \\
o1 & 10 & 0.8179 & 0.8600 & 0.8384 \\
o1 & 20 & 0.7962 & 0.8634 & 0.8285 \\
o1 & 50 & 0.7990 & 0.8947 & 0.8441 \\
o1 & 100 & 0.8107 & 0.8711 & 0.8398 \\
o1-mini & 10 & 0.7851 & 0.8757 & 0.8280 \\
o1-mini & 20 & 0.7632 & 0.8788 & 0.8169 \\
o1-mini & 50 & 0.7839 & 0.8905 & 0.8338 \\
o1-mini & 100 & 0.7900 & 0.8853 & 0.8350 \\
o3 & 10 & 0.8130 & 0.9341 & 0.8693 \\
o3 & 20 & 0.7843 & 0.9440 & 0.8568 \\
o3 & 50 & 0.7718 & 0.9521 & 0.8525 \\
o3 & 100 & 0.7976 & 0.9650 & 0.8734 \\
o3-mini & 10 & 0.8041 & 0.9017 & 0.8501 \\
o3-mini & 20 & 0.7938 & 0.9032 & 0.8450 \\
o3-mini & 50 & 0.7913 & 0.9094 & 0.8463 \\
o3-mini & 100 & 0.7965 & 0.9296 & 0.8579 \\
o3-mini & 200 & 0.7909 & 0.9263 & 0.8533 \\
o4-mini & 10 & 0.7769 & 0.9045 & 0.8359 \\
o4-mini & 20 & 0.7708 & 0.9162 & 0.8372 \\
o4-mini & 50 & 0.7690 & 0.9127 & 0.8347 \\
o4-mini & 100 & 0.7672 & 0.9428 & 0.8459 \\
sabia-3 & 10 & 0.6650 & 0.9320 & 0.7762 \\
sabia-3 & 20 & 0.6725 & 0.8944 & 0.7677 \\
sabia-3 & 50 & 0.6547 & 0.9446 & 0.7734 \\
sabia-3 & 100 & 0.7258 & 0.9968 & 0.8400 \\
sabiazinho-3 & 10 & 0.6958 & 0.9736 & 0.8116 \\
sabiazinho-3 & 20 & 0.6932 & 0.9737 & 0.8098 \\
sabiazinho-3 & 50 & 0.7163 & 0.9904 & 0.8313 \\
sabiazinho-3 & 100 & 0.6977 & 0.9868 & 0.8174 \\

\end{longtable}

\begin{longtable}{lrrrr}
\captionsetup{width=\linewidth}
\caption{Detailed results including precision and recall of each model in each top $k$ results considering only the cases in which there is a relevant result in the search results.} 
\label{tab:f1scoreidealdetailed}
\\
\toprule
Model & top $k$ & Precision & Recall & F1-score \\
\midrule
\endfirsthead
\toprule
Model & top $k$ & Precision & Recall & F1-score \\
\midrule
\endhead
\midrule
\multicolumn{5}{r}{Continues in the next page} \\
\midrule
\endfoot
\bottomrule
\endlastfoot

DeepSeek-R1 & 10 & 0.8189 & 0.8834 & 0.8499 \\
DeepSeek-R1 & 20 & 0.8074 & 0.9053 & 0.8536 \\
DeepSeek-R1 & 50 & 0.8068 & 0.9142 & 0.8571 \\
DeepSeek-R1 & 100 & 0.7923 & 0.9251 & 0.8536 \\
DeepSeek-R1 & 200 & 0.8101 & 0.9302 & 0.8660 \\
DeepSeek-R1-Distill-Qwen-1.5B & 10 & 0.5966 & 0.4733 & 0.5279 \\
DeepSeek-R1-Distill-Qwen-1.5B & 20 & 0.5531 & 0.4181 & 0.4762 \\
DeepSeek-R1-Distill-Qwen-1.5B & 50 & 0.5931 & 0.2507 & 0.3525 \\
DeepSeek-R1-Distill-Qwen-1.5B & 100 & 0.6364 & 0.0733 & 0.1315 \\
DeepSeek-R1-Distill-Qwen-7B & 10 & 0.5501 & 0.8712 & 0.6744 \\
DeepSeek-R1-Distill-Qwen-7B & 20 & 0.4776 & 0.8660 & 0.6156 \\
DeepSeek-R1-Distill-Qwen-7B & 50 & 0.5659 & 0.8939 & 0.6930 \\
DeepSeek-R1-Distill-Qwen-7B & 100 & 0.5755 & 0.8840 & 0.6972 \\
DeepSeek-V3 & 10 & 0.8196 & 0.9142 & 0.8643 \\
DeepSeek-V3 & 20 & 0.8125 & 0.9204 & 0.8631 \\
DeepSeek-V3 & 50 & 0.8159 & 0.9355 & 0.8716 \\
DeepSeek-V3 & 100 & 0.7955 & 0.9403 & 0.8618 \\
DeepSeek-V3 & 200 & 0.8085 & 0.9559 & 0.8760 \\
Llama-3-70B-Instruct & 10 & 0.7557 & 0.9738 & 0.8510 \\
Llama-3-70B-Instruct & 20 & 0.7663 & 0.9683 & 0.8555 \\
Llama-3-70B-Instruct & 50 & 0.7631 & 0.9684 & 0.8536 \\
Llama-3-70B-Instruct & 100 & 0.7556 & 0.9684 & 0.8488 \\
Llama-3.1-405B-Instruct & 10 & 0.8135 & 0.9345 & 0.8698 \\
Llama-3.1-405B-Instruct & 20 & 0.7924 & 0.9485 & 0.8634 \\
Llama-3.1-405B-Instruct & 50 & 0.7980 & 0.9341 & 0.8607 \\
Llama-3.1-405B-Instruct & 100 & 0.7970 & 0.9373 & 0.8615 \\
Llama-3.1-70B-Instruct & 10 & 0.7809 & 0.9558 & 0.8596 \\
Llama-3.1-70B-Instruct & 20 & 0.7806 & 0.9563 & 0.8596 \\
Llama-3.1-70B-Instruct & 50 & 0.7908 & 0.9451 & 0.8611 \\
Llama-3.1-70B-Instruct & 100 & 0.7758 & 0.9477 & 0.8532 \\
Llama-3.2-1B-Instruct & 10 & 0.2500 & 0.0025 & 0.0050 \\
Llama-3.2-1B-Instruct & 20 & 0.0000 & 0.0000 & 0.0000 \\
Llama-3.2-1B-Instruct & 50 & 0.5882 & 0.0257 & 0.0493 \\
Llama-3.2-1B-Instruct & 100 & 0.3200 & 0.0208 & 0.0391 \\
Llama-3.2-3B-Instruct & 10 & 0.6721 & 0.9018 & 0.7702 \\
Llama-3.2-3B-Instruct & 20 & 0.6973 & 0.8776 & 0.7771 \\
Llama-3.2-3B-Instruct & 50 & 0.6612 & 0.6547 & 0.6579 \\
Llama-3.2-3B-Instruct & 100 & 0.6770 & 0.8169 & 0.7404 \\
Llama-3.3-70B-Instruct & 10 & 0.8016 & 0.9164 & 0.8552 \\
Llama-3.3-70B-Instruct & 20 & 0.7964 & 0.9307 & 0.8583 \\
Llama-3.3-70B-Instruct & 50 & 0.8053 & 0.9217 & 0.8596 \\
Llama-3.3-70B-Instruct & 100 & 0.7984 & 0.9187 & 0.8543 \\
Llama-4-Maverick-Instruct-Basic & 10 & 0.7753 & 0.9654 & 0.8599 \\
Llama-4-Maverick-Instruct-Basic & 20 & 0.7830 & 0.9602 & 0.8626 \\
Llama-4-Maverick-Instruct-Basic & 50 & 0.7905 & 0.9520 & 0.8638 \\
Llama-4-Maverick-Instruct-Basic & 100 & 0.7961 & 0.9730 & 0.8757 \\
Llama-4-Scout-Instruct-Basic & 10 & 0.7848 & 0.9688 & 0.8671 \\
Llama-4-Scout-Instruct-Basic & 20 & 0.7854 & 0.9749 & 0.8699 \\
Llama-4-Scout-Instruct-Basic & 50 & 0.7748 & 0.9905 & 0.8694 \\
Llama-4-Scout-Instruct-Basic & 100 & 0.7641 & 0.9811 & 0.8591 \\
QwQ-32B & 10 & 0.7798 & 0.8882 & 0.8305 \\
QwQ-32B & 20 & 0.7743 & 0.9077 & 0.8357 \\
QwQ-32B & 50 & 0.7812 & 0.9303 & 0.8492 \\
QwQ-32B & 100 & 0.7747 & 0.9358 & 0.8476 \\
QwQ-32B & 200 & 0.7759 & 0.9663 & 0.8607 \\
gemini-2.0-flash & 10 & 0.7994 & 0.8466 & 0.8223 \\
gemini-2.0-flash & 20 & 0.7967 & 0.8698 & 0.8317 \\
gemini-2.0-flash & 50 & 0.7978 & 0.8446 & 0.8205 \\
gemini-2.0-flash & 100 & 0.7957 & 0.8732 & 0.8326 \\
gemini-2.0-flash-lite & 10 & 0.7702 & 0.9621 & 0.8555 \\
gemini-2.0-flash-lite & 20 & 0.7794 & 0.9569 & 0.8591 \\
gemini-2.0-flash-lite & 50 & 0.7816 & 0.9574 & 0.8607 \\
gemini-2.0-flash-lite & 100 & 0.7789 & 0.9665 & 0.8626 \\
gemini-2.0-pro-exp-02-05 & 10 & 0.8031 & 0.9281 & 0.8611 \\
gemini-2.0-pro-exp-02-05 & 20 & 0.7893 & 0.9453 & 0.8603 \\
gemini-2.0-pro-exp-02-05 & 50 & 0.8015 & 0.9696 & 0.8776 \\
gemini-2.0-pro-exp-02-05 & 100 & 0.7975 & 0.9522 & 0.8680 \\
gemini-2.0-pro-exp-02-05 & 200 & 0.7936 & 0.9613 & 0.8694 \\
gemini-2.5-pro-exp-03-25 & 10 & 0.8036 & 0.9284 & 0.8615 \\
gemini-2.5-pro-exp-03-25 & 20 & 0.7970 & 0.9486 & 0.8662 \\
gemini-2.5-pro-exp-03-25 & 50 & 0.7980 & 0.9634 & 0.8729 \\
gemini-2.5-pro-exp-03-25 & 100 & 0.8030 & 0.9499 & 0.8703 \\
gemini-2.5-pro-exp-03-25 & 200 & 0.7980 & 0.9643 & 0.8733 \\
gemma-2-27b-it & 10 & 0.7405 & 0.9732 & 0.8410 \\
gemma-2-27b-it & 20 & 0.7563 & 0.9869 & 0.8563 \\
gemma-2-27b-it & 50 & 0.7494 & 0.9837 & 0.8507 \\
gemma-2-27b-it & 100 & 0.7420 & 0.9773 & 0.8436 \\
gemma-3-27b-it & 10 & 0.7959 & 0.9750 & 0.8764 \\
gemma-3-27b-it & 20 & 0.7761 & 0.9652 & 0.8604 \\
gemma-3-27b-it & 50 & 0.7702 & 0.9683 & 0.8579 \\
gemma-3-27b-it & 100 & 0.7277 & 0.9800 & 0.8352 \\
gemma-3-4b-it & 10 & 0.7628 & 0.9042 & 0.8275 \\
gemma-3-4b-it & 20 & 0.7520 & 0.9135 & 0.8249 \\
gemma-3-4b-it & 50 & 0.7493 & 0.9073 & 0.8208 \\
gemma-3-4b-it & 100 & 0.7147 & 0.9145 & 0.8023 \\
gpt-4.1 & 10 & 0.8237 & 0.9125 & 0.8658 \\
gpt-4.1 & 20 & 0.8047 & 0.9169 & 0.8571 \\
gpt-4.1 & 50 & 0.8093 & 0.9290 & 0.8650 \\
gpt-4.1 & 100 & 0.8046 & 0.9233 & 0.8599 \\
gpt-4.1-mini & 10 & 0.8187 & 0.9433 & 0.8766 \\
gpt-4.1-mini & 20 & 0.8056 & 0.9608 & 0.8764 \\
gpt-4.1-mini & 50 & 0.7891 & 0.9666 & 0.8689 \\
gpt-4.1-mini & 100 & 0.7956 & 0.9642 & 0.8718 \\
gpt-4.1-nano & 10 & 0.8011 & 0.8657 & 0.8321 \\
gpt-4.1-nano & 20 & 0.7762 & 0.8278 & 0.8012 \\
gpt-4.1-nano & 50 & 0.7875 & 0.8348 & 0.8105 \\
gpt-4.1-nano & 100 & 0.7859 & 0.7976 & 0.7917 \\
gpt-4.5-preview & 10 & 0.8382 & 0.9133 & 0.8741 \\
gpt-4.5-preview & 20 & 0.8303 & 0.9244 & 0.8748 \\
gpt-4.5-preview & 50 & 0.8303 & 0.9390 & 0.8813 \\
gpt-4.5-preview & 100 & 0.8355 & 0.9420 & 0.8856 \\
gpt-4.5-preview & 200 & 0.8286 & 0.9391 & 0.8804 \\
gpt-4o & 10 & 0.8383 & 0.7195 & 0.7744 \\
gpt-4o & 20 & 0.8284 & 0.7171 & 0.7688 \\
gpt-4o & 50 & 0.8228 & 0.7919 & 0.8071 \\
gpt-4o & 100 & 0.8187 & 0.8069 & 0.8128 \\
gpt-4o-mini & 10 & 0.8037 & 0.9247 & 0.8599 \\
gpt-4o-mini & 20 & 0.7943 & 0.9307 & 0.8571 \\
gpt-4o-mini & 50 & 0.7903 & 0.9335 & 0.8560 \\
gpt-4o-mini & 100 & 0.7794 & 0.9453 & 0.8544 \\
o1 & 10 & 0.8479 & 0.8600 & 0.8539 \\
o1 & 20 & 0.8182 & 0.8634 & 0.8402 \\
o1 & 50 & 0.8182 & 0.8947 & 0.8547 \\
o1 & 100 & 0.8238 & 0.8711 & 0.8468 \\
o1-mini & 10 & 0.8199 & 0.8757 & 0.8469 \\
o1-mini & 20 & 0.7902 & 0.8788 & 0.8321 \\
o1-mini & 50 & 0.8091 & 0.8905 & 0.8479 \\
o1-mini & 100 & 0.8091 & 0.8853 & 0.8455 \\
o3 & 10 & 0.8534 & 0.9341 & 0.8919 \\
o3 & 20 & 0.8205 & 0.9440 & 0.8779 \\
o3 & 50 & 0.8030 & 0.9521 & 0.8712 \\
o3 & 100 & 0.8213 & 0.9650 & 0.8874 \\
o3-mini & 10 & 0.8365 & 0.9017 & 0.8679 \\
o3-mini & 20 & 0.8191 & 0.9032 & 0.8591 \\
o3-mini & 50 & 0.8141 & 0.9094 & 0.8591 \\
o3-mini & 100 & 0.8191 & 0.9296 & 0.8709 \\
o3-mini & 200 & 0.8072 & 0.9263 & 0.8626 \\
o4-mini & 10 & 0.8123 & 0.9045 & 0.8559 \\
o4-mini & 20 & 0.7969 & 0.9162 & 0.8524 \\
o4-mini & 50 & 0.7911 & 0.9127 & 0.8476 \\
o4-mini & 100 & 0.7825 & 0.9428 & 0.8552 \\
sabia-3 & 10 & 0.7154 & 0.9320 & 0.8095 \\
sabia-3 & 20 & 0.7169 & 0.8944 & 0.7959 \\
sabia-3 & 50 & 0.6964 & 0.9446 & 0.8018 \\
sabia-3 & 100 & 0.7627 & 0.9968 & 0.8642 \\
sabiazinho-3 & 10 & 0.7487 & 0.9736 & 0.8465 \\
sabiazinho-3 & 20 & 0.7363 & 0.9737 & 0.8385 \\
sabiazinho-3 & 50 & 0.7605 & 0.9904 & 0.8603 \\
sabiazinho-3 & 100 & 0.7299 & 0.9868 & 0.8392 \\
\end{longtable}

\begin{table}[H]
    \centering
    \caption{Llama-3.2-1B-Instruct model response for the query `acidente vascular cerebral' followed by the top 10 search resutls. K90 - Stroke/cerebrovascular accident is one of the relevant codes for this expression.}
    \begin{tabular}{p{13.5cm}}
    \toprule
    <answer selected\_code=``K90''></answer>\\
    \bottomrule
    \end{tabular}
    \label{tab:llama1example001}
\end{table}

\begin{table}[H]
    \centering
    \caption{Llama-3.2-1B-Instruct model response for the query `surdez' followed by the top 10 search resutls. H86 - Deafness is one of the relevant codes for this expression.}
    \begin{tabular}{p{13.5cm}}
    \toprule
    <h2>Surdez</h2>\\
    \\
    <p>Surdez é um termo que abrange uma série de condições que afetam a percepção auditiva, incluindo dificuldades de ouvir, audição distorcida, e outros sintomas relacionados à audição. Ela pode afetar pessoas de todas as idades e pode ser causada por uma variedade de fatores, incluindo doenças, lesões e condições médicas não relacionadas.</p>\\
    \\
    <p>De acordo com a Classificação Internacional de Códigos de Saúde (ICPC), a surdez pode ser classificada como uma condição de percepção auditiva (P24), que inclui surdez verbal, surdez mental e surdez histérica. Além disso, surdez de percepção (H86) também pode ser classificada como uma condição de percepção auditiva.</p>\\
    \\
    <p>Surdez de condução (H86) é uma forma mais específica de surdez de percepção que afeta as habilidades de condução de veículos. Surdez funcional (P75) é uma condição que afeta a capacidade de realizar atividades cotidianas, incluindo a audição.</p>\\
    \\
    <p>É importante notar que surdez pode ser um desafio para as pessoas, mas existem recursos e estratégias que podem ajudar a melhorar a qualidade de vida.</p>\\
    \\
    <answer selected\_code=``H86''>Surdez de percepção</answer>\\
    \bottomrule
    \end{tabular}
    \label{tab:llama1example002}
\end{table}

\begin{table}[H]
    \centering
    \caption{Llama-3.2-1B-Instruct model response for the query `inguinal' followed by the top 10 search results. In this case, due to the expression being too unspecific, the correct would be to not choose any code.}
    \begin{tabular}{p{13.5cm}}
    \toprule
    The result that best matches the query ``inguinal'' is: D06, dor inguinal\\
    \bottomrule
    \end{tabular}
    \label{tab:llama1example003}
\end{table}

\begin{table}[H]
  \caption{Spearman correlation tests, including the pair of variables assessed, $\rho$ and P-value.}
  \label{tab:correlation}
  \centering
  \begin{tabular}{lcc}
    \toprule
    Pair of variables & $\rho$     & P-value \\
    \midrule
    Max F1-score and model size \\
    \hspace{1cm}All models & 0.9084 & 0.0000\\
    \hspace{1cm}Models with 30B parameters or less & 0.8829 & 0.0084\\
    \hspace{1cm}Models bigger than 30B & 0.6810 & 0.0434\\
    Max F1-score and mean token usage per response & 0.4933 & 0.0035\\
    Max F1-score and mean price per response & 0.4599 & 0.0237 \\
    Max F1-score and mean time per response & 0.4321 & 0.0135 \\
    \bottomrule
  \end{tabular}
\end{table}

\newpage
\begin{table}
    \centering
    \caption{DeepSeek-R1 answer to the query `uso de droga' and a list of 200 results from the search engine.}
    \begin{tabular}{p{13.5cm}}
    \toprule
    <think> \\
    Okay, let's tackle this query. The user is asking for the ICPC code that corresponds to ``Uso de droga'' which translates to ``Drug use'' in English. The search results provided have several entries with codes P19 and A23. \\
    \\
    First, I need to understand the difference between A23 and P19. From the results, A23 seems to be associated with ``problema com o uso de drogas'' (problem with drug use) and ``uso de drogas como fator de risco'' (drug use as a risk factor). On the other hand, P19 entries are all about ``Abuso de drogas'' (drug abuse) and specific types of drug abuse like cocaine, cannabis, etc. \\
    \\
    The query is ``Uso de droga,'' which is a general term for drug use. If the user is referring to any drug use, even without specifying abuse, then A23 might be more appropriate because it's about problems related to drug use. However, if the context implies abuse, then P19 would be correct. Since the query is just ``use'' and not ``abuse,'' but in medical coding, sometimes ``use'' can be coded under abuse if it's problematic. But looking at the results, A23 explicitly mentions ``problema com o uso de drogas,'' which directly translates to ``problem with drug use,'' making it a better fit for general drug use issues. P19 is more about abuse or misuse. So unless the context specifies abuse, A23 is the right choice here. The third result under A23 is ``uso de medicamentos,'' which is medication use, but that's not the same as drug use in the illicit sense. The ninth result under A23 is ``uso de drogas como fator de risco,'' which again aligns with general drug use as a risk factor. Therefore, A23 seems to cover the general problem of drug use, whereas P19 is for abuse. The query doesn't mention abuse, so A23 is the best match.\\
    </think> \\
    \\
    <answer>A23</answer> \\
    \bottomrule
    \end{tabular}
    \label{tab:deepseekexample001}
\end{table}

\begin{longtable}{p{13.5cm}}
\caption{DeepSeek-R1's answer to the query `trombose venosa cerebral' (cerebral venous sinus thrombosis) and 200 results from the ICPC-2 search engine. The relevant codes in this case are K90 - Stroke/cerebrovascular accident and K91 - Cerebrovascular disease. The thesaurus, although, connects the expression `trombose dos seios cerebrovenoso' (cerebral venous sinus thrombosis) to the code N73 - Neurological infection other. Although infections may be related to this condition, the expression does not explicitly mention that and may lead to wrong coding.} 
\label{tab:deepseekexample003}
\\
\toprule
\endfirsthead
\toprule
\endhead
\midrule
\multicolumn{1}{r}{Continues in the next page} \\
\midrule
\endfoot
\bottomrule
\endlastfoot
<think>\\
Okay, let's tackle this query. The user is asking for the ICPC code for "trombose venosa cerebral," which translates to cerebral venous thrombosis. I need to look through the search results provided and find the best match.\\
\\
First, I'll scan the list of results for entries that specifically mention "venosa" and "cerebral." The query is about a venous thrombosis in the brain, so I need to focus on codes that specify venous involvement in the cerebral area.\\
\\
Looking at the results, I see several entries under code N73, like "trombose dos seios cerebrovenosos" and "trombose dos seios da face cerebral." The term "seios cerebrovenosos" refers to cerebral venous sinuses, which is exactly where a cerebral venous thrombosis would occur. Code N73 seems to be associated with thrombophlebitis and venous sinus thrombosis in the brain.\\
\\
There's also code W99 for cases during pregnancy, like "trombose venosa cerebral durante a gravidez," but the query doesn't mention pregnancy, so those can be ruled out. Similarly, W96 is for postpartum cases, which isn't relevant here.\\
\\
Another code to consider is K90, which has entries like "infarto cerebral em trombose venosa cerebral não-piogênica." This mentions cerebral infarction due to non-pyogenic venous thrombosis, which is a possible scenario. However, the primary condition here is the thrombosis itself, not the infarction complication.\\
\\
Code K91 has entries related to cerebral thrombosis but mostly arterial, like "trombose de uma artéria cerebral," which is arterial, not venous. The user's query specifies "venosa," so arterial codes are less likely.\\
\\
Code N73's entries like "trombose dos seios cerebrovenosos" directly refer to the venous sinuses in the brain, making it the most accurate match. Even though some entries under N73 mention thrombophlebitis, which is inflammation with a clot, cerebral venous thrombosis is a type of thrombophlebitis in the cerebral veins. Therefore, N73 is the correct code here.\\
\\
I should double-check if there's any other code that might be more specific. For example, K90's "infarto cerebral em trombose venosa cerebral não-piogênica" includes the thrombosis but is more about the resulting infarction. Since the query is about the thrombosis itself, N73 is more appropriate.\\
\\
So, the best code is N73.\\
</think>\\
\\
<answer>N73</answer>\\
\end{longtable}

\begin{longtable}{p{13.5cm}}
\caption{DeepSeek-R1's response to the query `m' and 20 results from the search engine.} 
\label{tab:deepseekexample002}
\\
\toprule
\endfirsthead
\toprule
\endhead
\midrule
\multicolumn{1}{r}{Continues in the next page} \\
\midrule
\endfoot
\bottomrule
\endlastfoot
<think>\\
Okay, let's tackle this query. The user input is just the letter `m'. The task is to find the best matching ICPC code from the provided search results.\\
\\
First, I need to understand what `m' could stand for in a medical context. Common possibilities might include `migraine', `myocardial infarction', `malaria', `melanoma', or abbreviations like `DM' (Diabetes Mellitus). But since the query is a single letter, it's a bit ambiguous.\\
\\
Looking at the search results, the codes and their expressions are:\\
\\
- T90: DM, DM2\\
- P73: mania\\
- A29: queda\\
- D96: mu-mu !\\
- T92: gota\\
- X10: X10\\
- A79: mesonefroma\\
- W85: DMG\\
- Y03: Y03\\
- Y16: Y16\\
- D99: mega-reto\\
- P06: mau sono\\
- Y02: Y02\\
- -34: ferro\\
- H15: H15\\
- Y25: Y25\\
- A78: uta\\
- R24: R24\\
- Y10: Y10\\
\\
The user's query is `m'. Let's see which of these expressions could start with or relate to `m'. \\
\\
Looking through the list:\\
- `DM' and `DM2' (T90) stand for Diabetes Mellitus, which is often abbreviated as DM. But `m' alone might not be enough to map to DM. However, if the user intended `DM', then T90 would be the code. But the query is `m', not `dm'.
- `mania' (P73) starts with `m', so that's a possibility. Mania is a mental health condition.\\
- `mu-mu !' (D96) has `mu' but not sure if that's related.\\
- `mega-reto' (D99) starts with `m' as well. `Mega' could be a prefix, but `mega-reto' isn't a standard term I recognize.\\
- `mau sono' (P06) starts with `m', but `mau' might be a term in another language. In Portuguese, `mau' means `bad', but `mau sono' could translate to `bad sleep', which might relate to insomnia. However, the code P06 isn't familiar to me for that.\\
- `mesonefroma' (A79) starts with `m', but I'm not sure what that term refers to. It might be a misspelling or a less common condition.\\
\\
Other entries like Y03, Y16, etc., are just codes with the same letters, which don't seem relevant. The code `-34' for `ferro' (iron) is a negative number, which might be an error.\\
\\
Considering the possible matches, `mania' (P73) starts with `m' and is a recognized medical condition. `DM' (T90) is Diabetes Mellitus, but the query is `m', not `dm'. However, in some contexts, `m' could be a typo for `dm', but that's speculative. `Mania' seems more directly related to the letter `m'. However, I need to check if there's a more appropriate code. \\
\\
Wait, `mau sono' (P06) is another possibility. If `mau' is part of the term, but the query is just `m', maybe the user intended the first letter. However, `mau' in Portuguese is `bad', so `mau sono' might be `insomnia' or sleep issues, but the code P06 isn't standard for that. \\
\\
Alternatively, `mesonefroma' (A79) is unclear. Maybe a tumor type? Not sure.\\ 
\\
The code D96: `mu-mu !' is unclear. Maybe a typo or slang. \\
\\
So between `mania' (P73) and `DM' (T90), which is more likely? If the user typed `m' intending `dm', then T90. But if they meant a condition starting with `m', then P73. Since the query is a single letter, it's ambiguous, but `mania' starts with `m', so P73 might be the best fit. However, in medical coding, abbreviations are common. For example, DM is a standard abbreviation for Diabetes Mellitus. But the query is `m', not `dm'. Unless `m' is part of a longer abbreviation, but that's a stretch. \\
\\
Alternatively, maybe the user made a typo and meant `dm', but without more context, it's hard to say. The instructions say to pick the best match. If `m' is the first letter, then `mania' (P73) is the best. However, in some coding systems, `M' might refer to musculoskeletal issues, but in ICPC, the chapters are different. \\
\\
Wait, checking ICPC structure: The first letter of the code represents the chapter. For example, T is endocrine, P is psychological. So `m' as a query might not correspond to the chapter. The user's query is the letter `m', but the codes are in the format like T90, P73, etc. So maybe the user is referring to a chapter, but ICPC codes are alphanumeric with the first letter as the chapter. However, the search results don't have any code starting with M. The chapters in ICPC are A (General), B (Blood), D (Digestive), etc. There's no M chapter. So perhaps the user is referring to a term starting with M. \\
\\
In that case, the best matches are `mania' (P73), `mesonefroma' (A79), `mega-reto' (D99), `mau sono' (P06). Among these, `mania' is a well-known condition, so P73 is the most likely correct code. The others are either unclear or not standard terms. \\
\\
Therefore, the best answer is P73.\\
</think>\\
\\
<answer>P73</answer>\\
\end{longtable}

\begin{landscape}
\vfill
\begin{figure}
\centering
\includegraphics[scale=0.48]{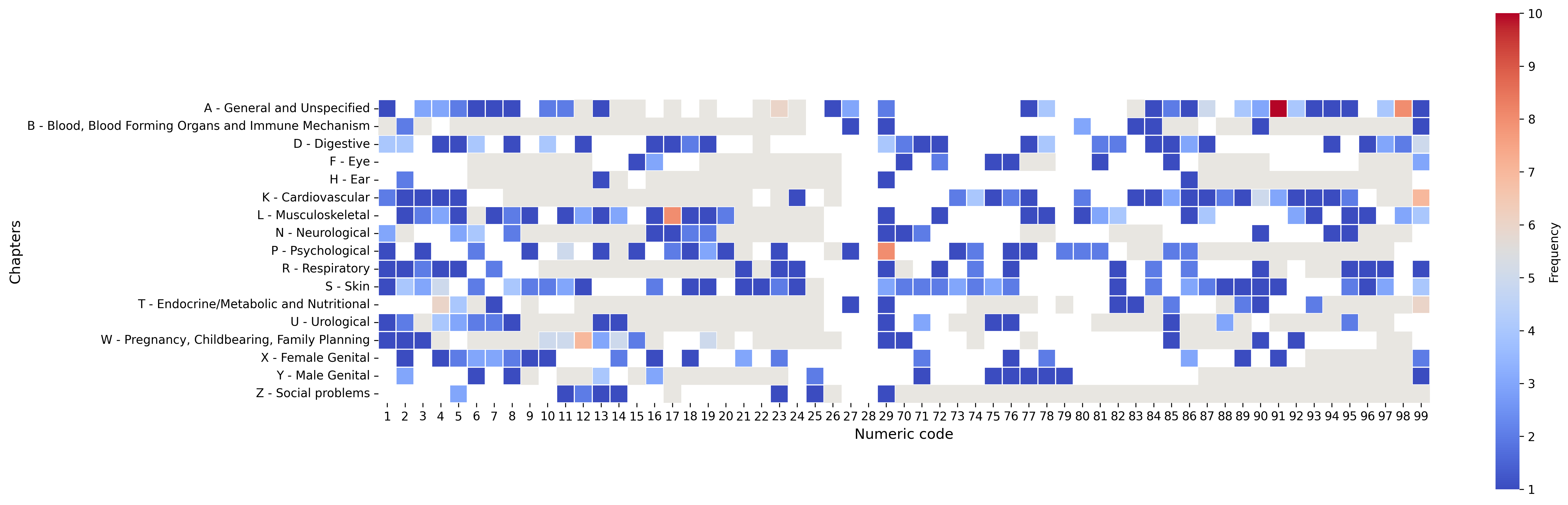}
\caption{Frequency map of ICPC-2 codes in the evaluation dataset (except process codes). A blank square means that the corresponding code is not represented in the evaluation dataset. A gray square means that the corresponding code does not exist in ICPC-2.}
\label{fig:heatmap_1}
\end{figure}

\vfill
\begin{figure}
\centering
\includegraphics[scale=0.8]{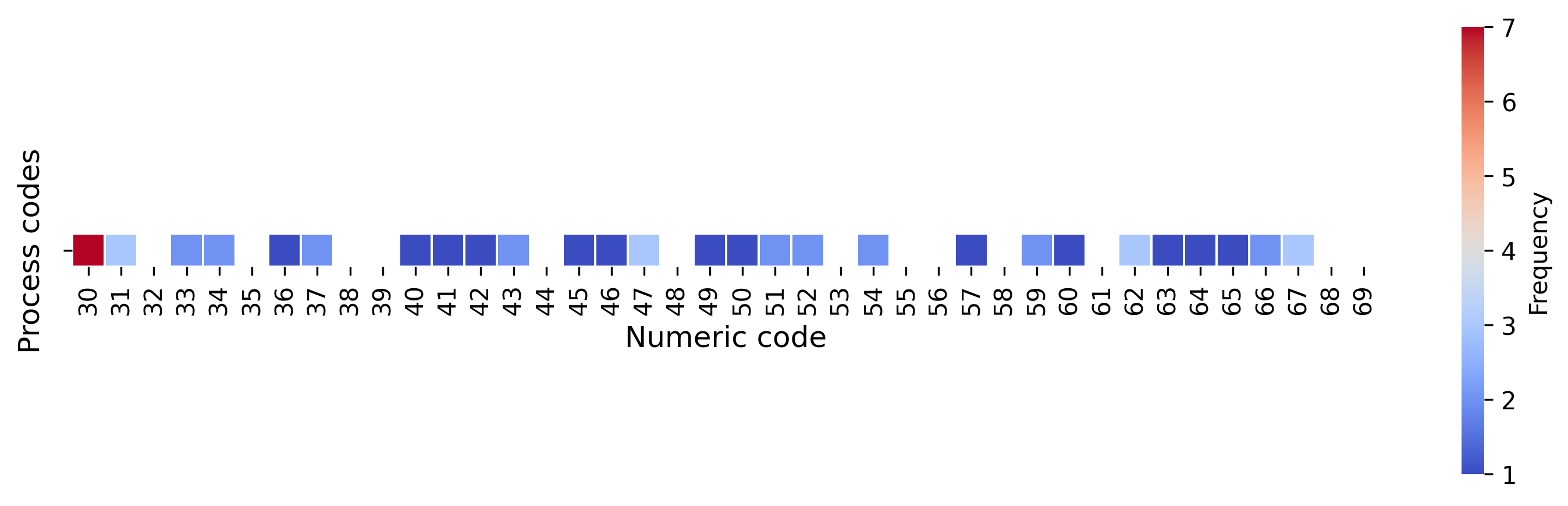}
\caption{Frequency map of process ICPC-2 codes in the evaluation dataset. A blank square means that the corresponding code is not represented in the evaluation dataset. A gray square means that the corresponding code does not exist in ICPC-2.}
\label{fig:heatmap_2}
\end{figure}
\vfill
\end{landscape}

\begin{landscape}
\vfill
\begin{figure}
\centering
\includegraphics[scale=0.6]{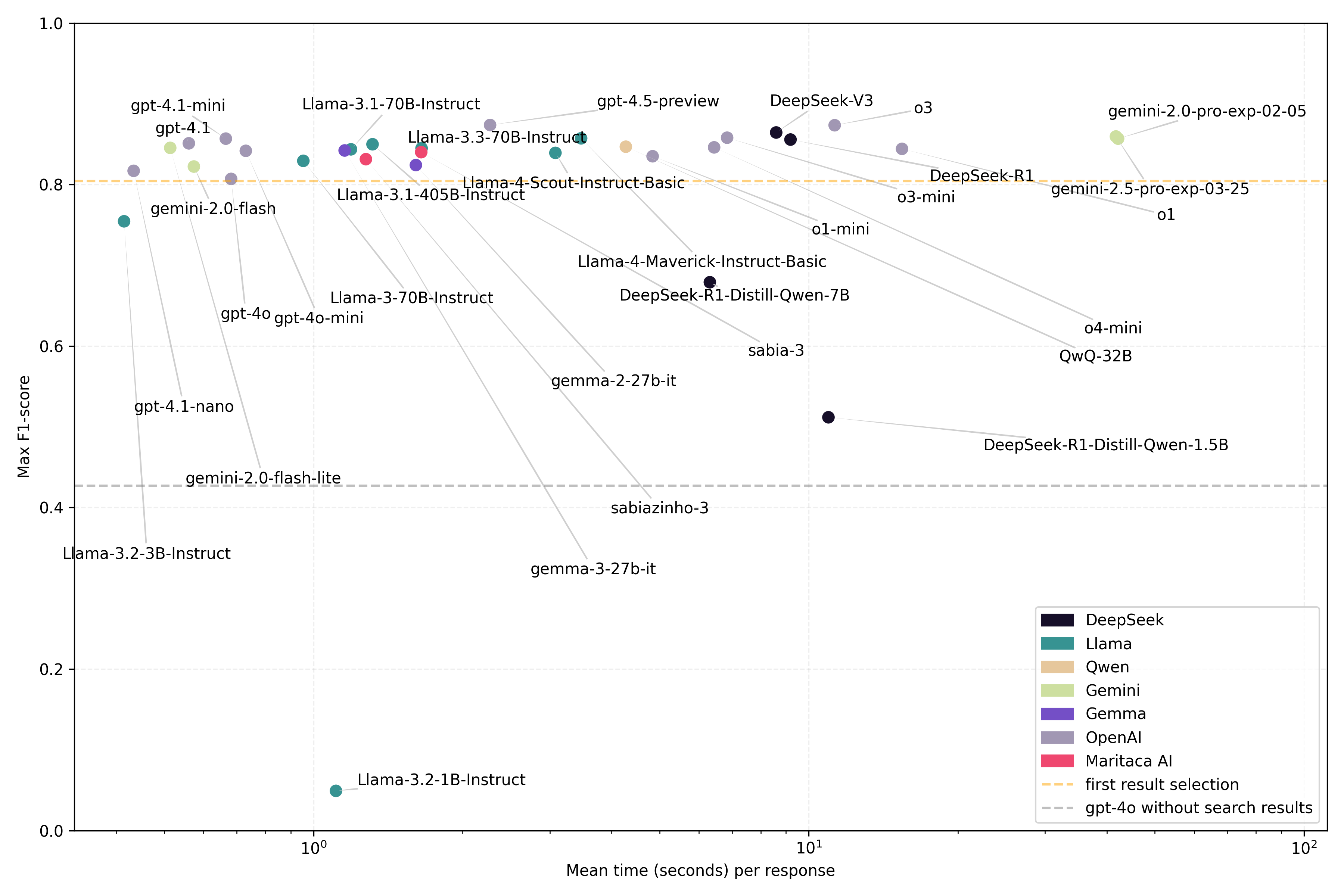}
\caption{Relationship between mean time per response in seconds and F1-score. For each model, the max F1-score was considered. Note the x-axis in log scale. The time estimate for the models gemini-2.0-pro-exp-02-05 and gemini-2.5-pro-exp-03-25 are probably overestimated due to rate limiting of experimental models in Google's API.}
\label{fig:timevsf1}
\end{figure}
\vfill
\end{landscape}

\begin{landscape}
\vfill
\begin{figure}
\vspace{1cm}
\centering
\includegraphics[scale=0.6]{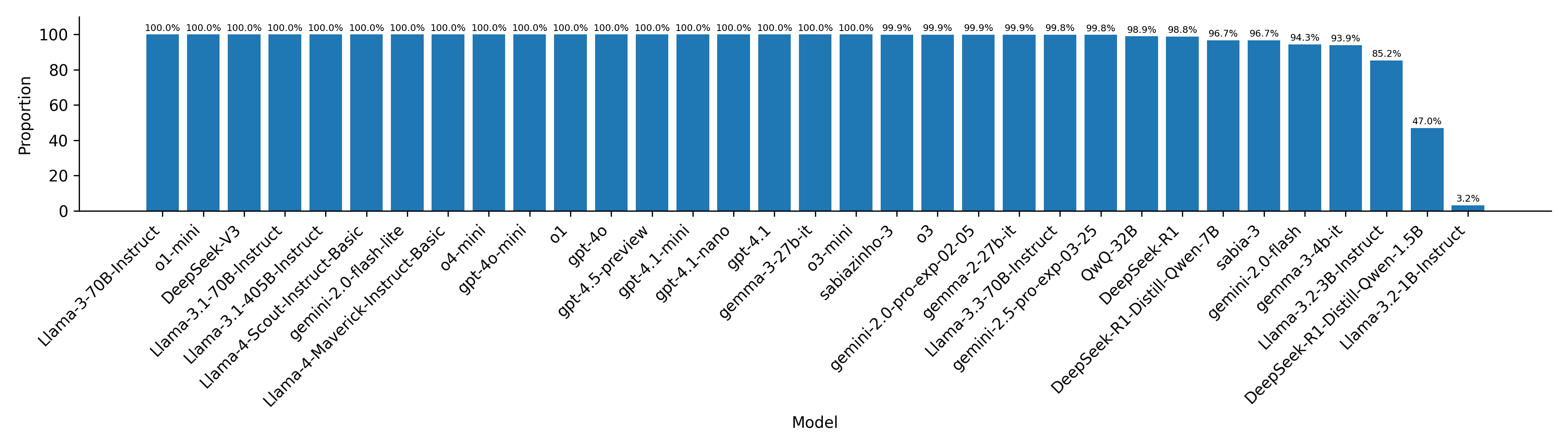}
\caption{Proportion of responses that followed the format specified in the response.}
\label{fig:rightformatanswer}
\end{figure}

\vfill
\begin{figure}
\vspace{1cm}
\centering
\includegraphics[scale=0.6]{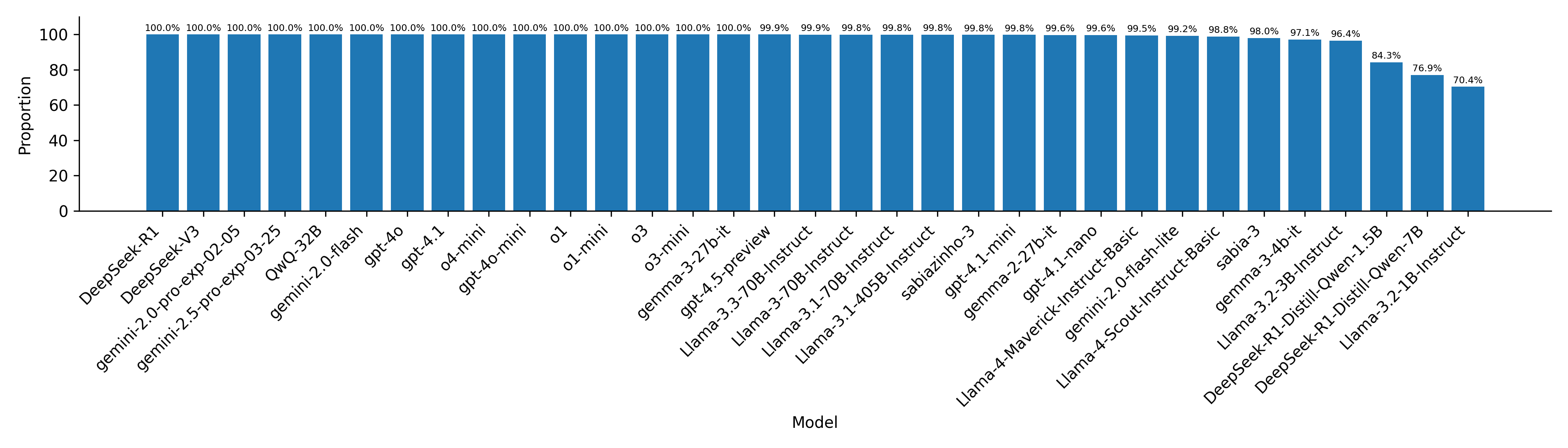}
\caption{Proportion of responses with a valid format in which the selected code was a valid ICPC-2 code.}
\label{fig:validicpcgraph}
\end{figure}

\vfill
\begin{figure}
\vspace{1cm}
\centering
\includegraphics[scale=0.60]{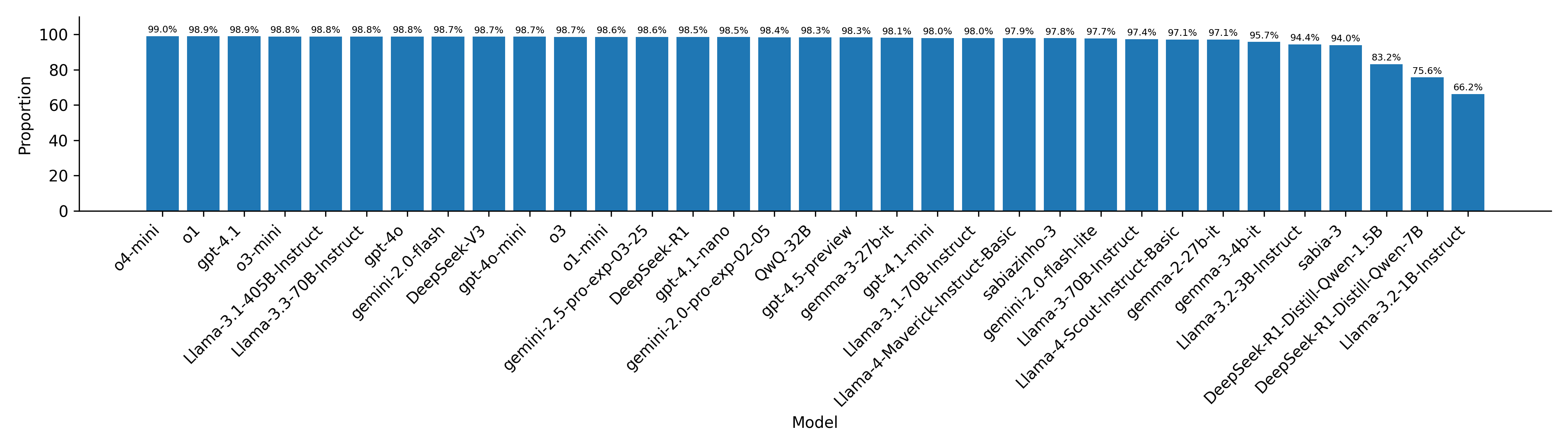}
\caption{Proportion of responses with a valid format in which the selected code was valid and present among the search engine results.}
\label{fig:selectedcodeinresults}
\end{figure}
\vfill
\end{landscape}

\begin{landscape}
\vfill
\begin{figure}
\centering
\includegraphics[scale=0.6]{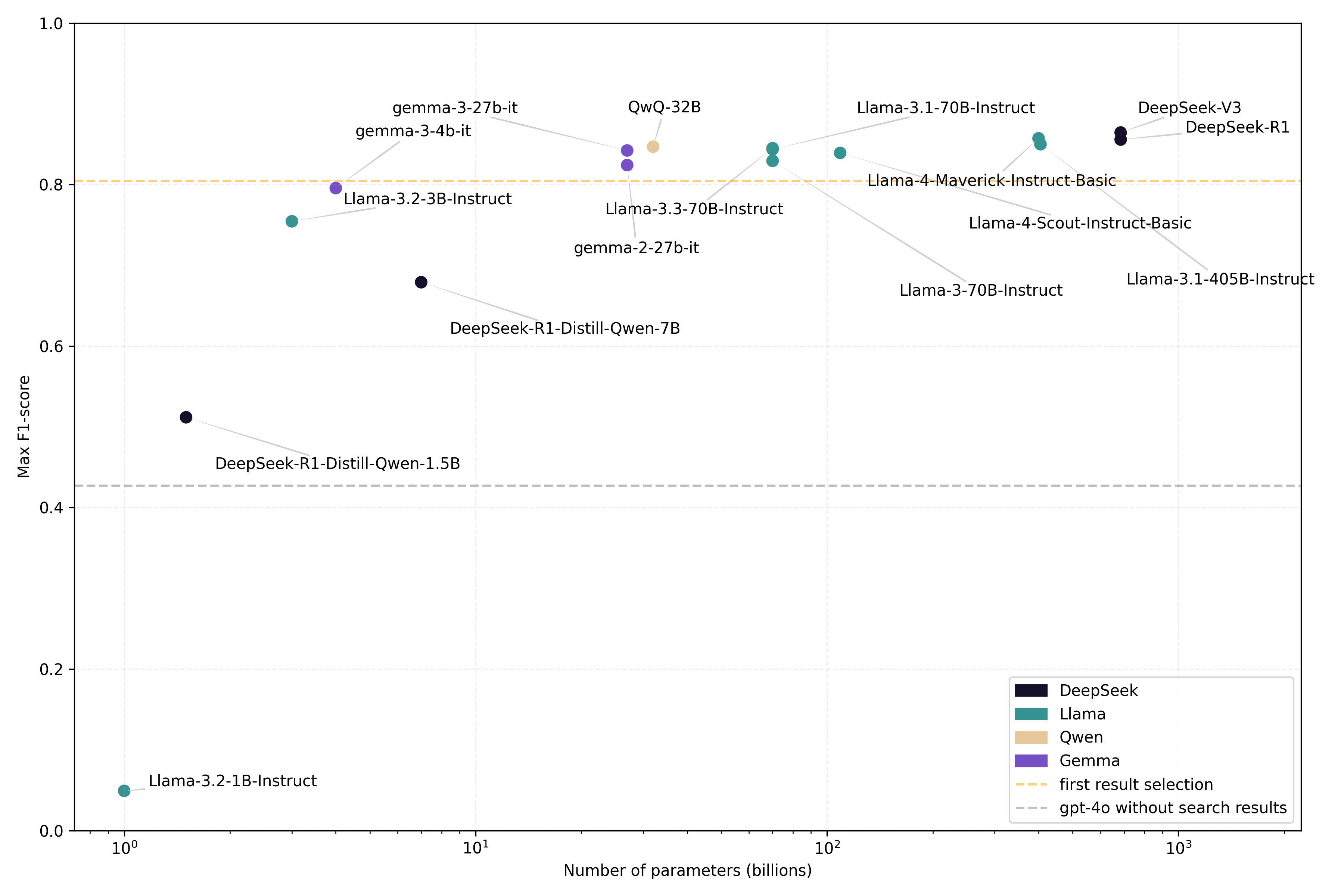}
\caption{Relationship between model's number of parameters and F1-score. For each model, the max F1-score was considered. Note the x-axis in log scale. The number of parameters is expressed in billions of parameters. Only open-source models were included.}
\label{fig:nparamsf1}
\end{figure}
\vfill
\end{landscape}

\end{document}